\documentclass[journal]{IEEEtran}

\usepackage{multirow}
\usepackage{multicol}
\usepackage{amsthm}
\usepackage{makecell}
\usepackage{graphicx}
\usepackage[caption=false]{subfig}
\usepackage{pifont}
\usepackage{amsmath,amsfonts}
\usepackage{algorithmic}
\usepackage{array}
\usepackage{textcomp}
\usepackage{stfloats}
\usepackage{url}
\usepackage{verbatim}
\usepackage{balance}
\usepackage{cite}
\usepackage{color}
\usepackage{booktabs}
\usepackage[colorlinks]{hyperref}
\hypersetup{
    linkcolor=blue,
    filecolor=blue,      
    urlcolor=blue,
    citecolor=blue,}

\makeatletter
\renewcommand{\@cite}[1]{\textcolor{blue}{\text{[#1]}}}
\makeatother

\newcommand{\cmark}{\ding{51}}%
\newcommand{\xmark}{\ding{55}}%
\newtheorem{defn}{Definition}

\begin{document}
\title{CityNet: A Comprehensive Multi-Modal Urban Dataset for Advanced Research in Urban Computing}
\author{Zhengfei Zheng, Xu Geng, and Hai Yang
\thanks{This work was supported by the Hong Kong Research Grants Council under a Theme-based Research Scheme, with Grant T41-603/20R. (\emph{Corresponding author: Xu Geng.})}
\thanks{Zhengfei Zheng and Hai Yang are with the Department of Civil and Environmental Engineering, the Hong Kong University of Science and Technology
Clear Water Bay, Kowloon, Hong Kong. (E-mail: zzhengak@connect.ust.hk; cehyang@ust.hk)}
\thanks{Xu Geng is with the Department of Computer Science and Engineering,
the Hong Kong University of Science and Technology, Clear Water Bay, Kowloon, Hong Kong. (E-mail: xgeng@connect.ust.hk)}}


\maketitle

\begin{abstract}
Data-driven approaches have emerged as a popular tool for addressing challenges in urban computing. However, current research efforts have primarily focused on limited data sources, which fail to capture the complexity of urban data arising from multiple entities and their interconnections. Therefore, a comprehensive and multifaceted dataset is required to enable more extensive studies in urban computing. In this paper, we present CityNet, a multi-modal urban dataset that incorporates various data, including taxi trajectory, traffic speed, point of interest (POI), road network, wind, rain, temperature, and more, from seven cities. We categorize this comprehensive data into three streams: mobility data, geographical data, and meteorological data. We begin by detailing the generation process and basic properties of CityNet. Additionally, we conduct extensive data mining and machine learning experiments, including spatio-temporal predictions, transfer learning, and reinforcement learning, to facilitate the use of CityNet. Our experimental results provide benchmarks for various tasks and methods, and also reveal internal correlations among cities and tasks within CityNet that can be leveraged to improve spatiotemporal forecasting performance. Based on our benchmarking results and the correlations uncovered, we believe that CityNet can significantly contribute to the field of urban computing by enabling research on advanced topics.
\end{abstract}


\section{Introduction}

\IEEEPARstart{D}{ata-driven} analytical techniques have become increasingly prevalent in both the research community and industry for addressing various tasks in urban computing \cite{zheng2014urban}. In recent years, several machine learning techniques, including deep learning \cite{zhang2016dnn,li2017diffusion}, transfer learning \cite{wang2019cross,yao2019learning}, and reinforcement learning \cite{xu2018reinforcement,li2019efficient}, have been proposed to solve urban management problems such as predicting traffic speeds and flows, managing ride-hailing fleets, and forecasting urban environments. Despite the growing interest in urban computing, the absence of an ideal open-source dataset has restricted research efforts to individual tasks using data from limited sources. Without multi-modal dataset, these approaches fall short of building a smart city system that can leverage data from multiple sources to make more intelligent decisions. For instance, while most research on taxi fleet management systems focuses solely on taxi data, such a system could benefit from a range of additional data sources, such as real-time traffic speeds, weather conditions, point-of-interest distributions, and other modes of transportation's demands and supplies, all of which are closely related to taxi demands. Therefore, it is crucial to develop a comprehensive and multifaceted dataset that can facilitate more extensive studies in urban computing and enable the development of smart city systems.

Regrettably, currently available open datasets, such as PeMS \cite{chen2001freeway}, METR \cite{H2014Big} and NYC Cabs \cite{ferreira2013visual} are limited to either traffic speeds or taxi-related data. Consequently, they do not provide inclusive support for studies on a realistic and comprehensive smart city system. Moreover, individual datasets cannot be easily merged into an all-encompassing dataset due to variations in their temporal ranges. For example, while some datasets such as TaxiBJ \cite{zhang2016dnn}, T-Drive \cite{yuan2011driving} and Q-Traffic \cite{bbliaojqZhangKDD18deep} all pertain to Beijing, they are not temporally aligned and thus cannot be jointly utilized. This limitation underscores the urgent need for a comprehensive and spatio-temporally aligned dataset in urban computing to facilitate more precise algorithms and insightful analyses. Such a dataset would enable researchers to study the complexities of urban data arising from multiple entities and their interconnections, thus supporting the development of more effective smart city systems.

In brief, the creation and implementation of a comprehensive urban dataset encounter two major challenges. Firstly, urban data are usually fragmented across different entities, such as governmental bodies and private enterprises, resulting in disparities in data acquisition and processing protocols. These differences may manifest in variations in spatio-temporal coverage, granularity, and attributes. Consequently, integrating these disparate data sources into a standardized format with aligned range for research purposes poses a significant hurdle. Secondly, beyond data collection, identifying interdependencies among various datasets is critical to enhance performance by sharing and transferring relevant knowledge. Therefore, uncovering these interdependencies is crucial to improve the overall effectiveness of the dataset.

\begin{figure*}[ht]
    \centering
    \includegraphics[width=\linewidth]{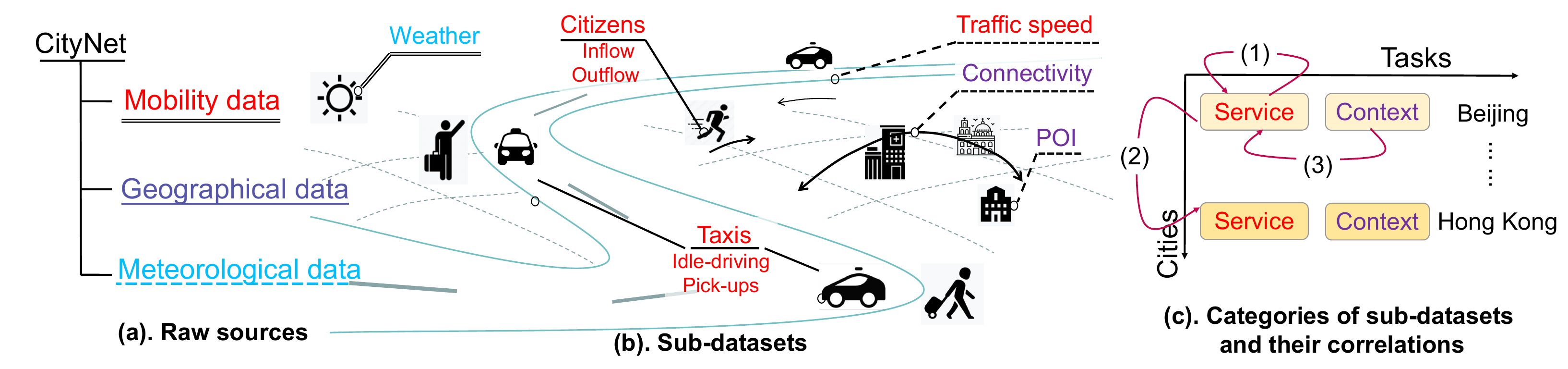}
    \vspace{-0.3cm}
    \caption{Architecture of CityNet.\textit{Left}: Three raw data sources of CityNet. \textit{Middle}: Schematic description of all 8 sub-datasets, whose sources are distinguished by color as shown in Fig. 1(a) and  1(b). \textit{Right}: Decomposition of the data dimensions into cities and tasks. Directed curves indicate correlations to be discovered in this paper. \centering}
    \label{fig:architecture}
    \vspace{-0.4cm}
\end{figure*}

In order to address above challenges, this paper introduces CityNet, a multi-modal dataset comprising data from various cities and sources for smart city applications. Drawing inspiration from \cite{wei2016transfer}, we use the term ``multi-modal" to reflect the diverse range of cities and sources from which CityNet is derived. In comparison to existing urban datasets, CityNet offers the following distinctive features:

\begin{itemize}
    \item \textbf{Comprehensiveness}: Fig. \ref{fig:architecture}(a), illustrates that CityNet comprises three types of raw data (mobility data, geographical data, and meteorological data) collected from seven different cities. Furthermore, we have processed the raw data into several sub-datasets (as shown in Fig. \ref{fig:architecture}(b)) to to capture a wider range of urban phenomena. For instance, we have transformed raw mobility data of taxi movements into region-based measurements such as taxi flows, pickups, and idle driving time. These measurements are crucial in revealing the state of the transportation market and citizen activities.
    
    \item \textbf{Spatio-temporal Alignment}: In order to facilitate the analysis of correlations between cities and tasks, we have adopted a consistent spatio-temporal configuration for all sub-datasets from the same city. This includes a uniform range of geographical coordinates, temporal frames, and sampling intervals.
    
    \item \textbf{Interrelationship}: 
        We have classified the sub-datasets into two categories: service data and context data, as depicted in Fig. \ref{fig:architecture}(c). Service data pertains to the status of urban service providers (e.g. taxi companies), while context data refers to the urban environment (e.g. weather). Based on this categorization, we have formulated and tested three types of correlations, as shown in Fig. \ref{fig:architecture}(c), correlations (1) among mobility services, (2) among context, such as urban geography, and (3) between contexts and services. 
\end{itemize}

CityNet's comprehensive and correlated data make it a valuable resource for machine learning tasks in urban computing. These tasks include spatio-temporal predictions and its multi-task variant, spatio-temporal transfer learning, and reinforcement learning. In this paper, we present extensive benchmarking results for these tasks.

The contributions of this paper are summarized as follows:
\begin{itemize}
    \item To the best of our knowledge, CityNet is \textbf{the first multi-modal urban dataset} that aggregates and aligns sub-datasets from various tasks and cities. Using CityNet, we have provided a wide range of benchmarking results to inspire further research in areas such as spatio-temporal predictions, transfer learning, reinforcement learning, and federated learning in the field of urban computing.
    
    \item Our analyses and experiments on CityNet have yielded valuable insights for researchers. Our studies have confirmed the \textbf{correlations} among sub-datasets and have demonstrated that urban modeling and analyses can be enhanced by appropriately utilizing the mutual knowledge among correlated sub-datasets. To this end, we have employed data mining tools to uncover the correlations between service and context data, and have utilized machine learning results to showcase the correlations among cities and tasks.. 
\end{itemize}

The paper is structured as follows. Section \ref{sec:architecture} outlines the pre-processing procedure of all sub-datasets in CityNet, along with their basic statistics. In Section \ref{sec:mining}, we employ data mining tools to reveal and elucidate the correlations between contexts and service data. In Section \ref{sec:apps}, we conduct machine learning experiments, such as spatio-temporal predictions, transfer learning, and reinforcement learning on CityNet, to provide comprehensive benchmarking results and to demonstrate the interrelationships among services and cities.

\section{Architecture of CityNet}
\label{sec:architecture}
    

\begin{table*}[ht]
\caption{Overview of data in CityNet, including spatio-temporal ranges and availability of sub-datasets in each city.}
\vspace{-0.2cm}
\resizebox{\linewidth}{!}{\begin{tabular}{cc|cccc|cccc|c}
    \toprule
    \multirow{2}[0]{*}{City} & \multirow{2}[0]{*}{\makecell{Time Span\\(inclusive)}}& \multicolumn{4}{c|}{Geographical data} & \multicolumn{4}{c|}{Mobility data} & \multirow{2}[0]{*}{Meteorology} \\
    
     &  & Longitude range (E) & Latitude range (N) & POI & Road & Speed & In/Outflow & Pickup & Idle\\
     \midrule
    Beijing  & 10/20 - 10/29, 2013 & 115.9045-116.9565 & 39.5786-40.4032 & \cmark & \cmark & \xmark & \cmark & \cmark & \cmark & \cmark\\
    Shanghai & 08/01 - 08/31, 2016 & 121.2054-121.8176 & 30.9147-31.4202 & \cmark & \cmark & \xmark & \cmark & \cmark & \cmark & \cmark\\
    Shenzhen & 12/01 - 12/31, 2015 & 113.8097-114.3087 & 22.5085-22.7739 & \cmark & \cmark & \xmark & \cmark & \cmark & \cmark & \cmark\\
    Chongqing & 08/01 - 08/31, 2017& 106.3422-106.6486 & 29.3795-29.8322 & \cmark & \cmark & \xmark  & \cmark & \cmark & \cmark & \cmark \\
    Xi‘an & 10/01 - 11/30, 2016 & 108.9219-109.0093 & 34.2049-34.2799& \cmark & \cmark & \cmark  & \cmark & \cmark & \xmark & \cmark\\
    Chengdu & 10/01 - 11/30, 2016  & 104.0421-104.1296 & 30.6528-30.7278 & \cmark & \cmark & \cmark & \cmark & \cmark & \xmark & \cmark\\
    Hong Kong & 10/13 - 12/22, 2020  & 113.9878-114.2459 & 22.2745-22.4670 & \xmark & \cmark & \cmark & \multicolumn{3}{c|}{\xmark} & \cmark\\
    \bottomrule
\end{tabular}}
\label{tab:citynet_overview}
\vspace{-0.4cm}
\end{table*}
This section provides an overview of the raw data formats, pre-processing procedures, and processed sub-datasets of CityNet.

\subsection{Overview}
The data structure for a single city in CityNet is presented in Fig. \ref{fig:architecture}. The figure showcases the three types of raw data that are collected from each city, namely: 
\begin{itemize}
    \item \textbf{Mobility data}: The mobility data in CityNet primarily consists of taxi movements, which provide valuable insights into citizen activities and the state of the transportation network. For instance, region-wise taxi flows can reveal urban crowd movement patterns, while taxi pickup and idle driving data can serve as proxies for transport demand and supply, respectively. To this end, we collect raw taxi GPS points from each city and process them into sub-datasets, including inflow  ($\mathbf{x}^{in}$), outflow ($\mathbf{x}^{out}$), pickup ($\mathbf{x}^{pickup}$) and idle driving ($\mathbf{x}^{idle}$). Furthermore, we gather traffic speed ($\mathbf{x}^{speed}$) data  to provide a real-time depiction of the transportation system's status.
    
    \item \textbf{Geographical data}: In order to depict the comprehensive layout of the city, we collect data pertaining to Point-of-Interest (POI) and road network. The POI data ($\mathbf{x}^{poi}$) provides information on the purpose and utility of each area, whereas the road network data provides insight into the connectivity and accessibility among POIs. Additionally, we generate a region-wise adjacency matrix ($\mathbf{A}^{conn}$) to explicate the ease of travel among different regions.
    
    \item \textbf{Meteorological data}: Meteorological phenomena, such as rainfall or low temperatures, can have a profound impact on human activities. For instance, during a rainy day, the demand for taxis may increase while urban flows may decrease. As a result, we accumulate meteorological data ($\mathbf{x}^{mtr}$) to account for these factors. 
\end{itemize}

 In order to facilitate a clear understanding of the data used in this study, we have classified all taxi-related mobility data (including flow, pickup, and idle driving and traffic speed data) as \emph{service data}, as they pertain to the operational states of transport service providers. Accordingly, all other data have been categorized as \emph{context data}. Furthermore, we have provided concise definitions of the key concepts employed throughout this paper.


\begin{defn}[Taxi GPS Points]
The GPS points of taxis in a given city, denoted by $\mathcal{G}_c$, comprise the trajectories of taxis $\mathcal{G}_c = \{tr_a\}_{a\in \mathcal{A}_c}$, where $\mathcal{A}_c$ represents the set of taxis operating in a city $c$. Each trajectory $tr_a$ is a sequence of tuples $tr_a = \left[\left(pos_k, t_k, occ_k\right)\right]_{k=1}^{|tr_a|}$, where $pos_k$ indicates the position (i.e. longitude and latitude) of the taxi at time $t_k$, and $occ_k$ denotes the occupancy status of the taxi at time at time $t_k$. 
\end{defn}

\begin{defn}[POI]
A point-of-interest (POI) can be defined as $poi = (pos_{poi}, cat_{poi})$, where $pos_{poi}$ represents the position of the POI (i.e., longitude and latitude), and $cat_{poi}$ indicates the category of the POI (e.g., catering, shopping). The set of all POIs in a city $c$ id denoted as $POI_c$.
\end{defn}

\begin{defn}[Road segments and real-time speed]
A road segment can be defined by its two endpoints as $seg = (pos_{0},pos_{1})$. Each segment is linked to its historical real-time traffic speeds, denoted as $sp_{seg} =[(t_{k},speed_{k})]_{k=1}^{|sp_{seg}|}$, where $speed_{k}$ denotes the speed of $seg$ at time $t_{k}$.
\end{defn}


\begin{defn}[Region]
Each city $c$ is split into $W_c\times H_c$ square grids, each with size 1km$\times$1km. Each grid in a city is referred to as a \textbf{region}, denoted as $r$. We denote the set of all regions in city $c$ as $\mathcal{R}_c$, and use $r_c(i, j)$ to denote the region in city $c$ with coordinate $(i, j)$. 
\end{defn}

\begin{defn}[Timestamps]
We define the set of available time stamps of city $c$ as $\mathcal{T}_c = [\tau_c - T_c + 1, ...\tau_c]$, where $\tau_c$ denotes the last timestamp, and $T_c$ is the number of available time stamps of city $c$. By default, we choose each time stamp to be a period of 30 minutes. 
\end{defn}

\begin{defn}[Spatio-temporal Tensors]
We denote a spatio-temporal tensor for city $c$ (e.g. taxi flow values) as $\mathbf{x}_c\in \mathbb{R}^{T_c \times W_c\times H_c}$, and use $\mathbf{x}_c(\tau, i, j)$ and $\mathbf{x}_c(\tau)$ to denote the value at region $r_c(i, j)$ and the values for all regions in city $c$ at timestamp $\tau$, respectively. 
\end{defn}


Our data collection covers a total of 7 cities, namely Beijing, Shanghai, Shenzhen, Chongqing, Xi'an\footnotemark, Chengdu\footnotemark[\value{footnote}] and Hong Kong. 
Table \ref{tab:citynet_overview} provides details on the properties of the collected data, including data range, size, and availability. It is important to note that due to limitations in data availability, not all types of data are accessible for each city. For ease of reference, we have compiled a list of notations used in this paper in Table \ref{tab:notation} in the Appendix.

\footnotetext{The original data were obtained from the HKUST-DiDi Joint Research Laboratory. Some of the data can be made available upon request after undergoing a process of desensitization.}

\begin{table*}[t]
\centering
\caption{The sub-datasets of taxi mobility, road connectivity, and traffic speed for all cities are described by their respective statistical features. In particular, the temporal granularity for in/outflow data is 30 minutes, while pickup, idle driving, and traffic speed data are recorded at 10-minute intervals. The degree of sparsity, defined as the fraction of zero entries within the sub-dataset, is also reported to provide insight into the completeness and quality of the data.}
\vspace{-0.2cm}
\begin{tabular}{c|ccc|ccc|ccc}
\toprule
\multirow{2}[0]{*}{Taxi mobility}& In/Outflow   & Pickup         & Idle-time        & In/Outflow   & Pickup        & Idle-time        & In/Outflow   & Pickup        & Idle-time       \\ \cmidrule{2-10}
        & \multicolumn{3}{c|}{Beijing}                      & \multicolumn{3}{c|}{Shanghai}                    & \multicolumn{3}{c}{Shenzhen}                    \\ \midrule
Shape   & (480,106,92) & \multicolumn{2}{c|}{(1440,106,92)} & (1448,62,57) & \multicolumn{2}{c|}{(4464,62,57)} & (1448,50,30) & \multicolumn{2}{c}{(4464,50,30)} \\
Mean    & 6.682        & 0.389           & 0.488            & 27.181       & 0.637          & 1.618            & 88.244       & 2.893          & 5.488           \\
Max     & 671.0        & 340.0           & 243.3            & 1440.0       & 304.0          & 438.0            & 3601.0       & 302.0          & 583.7           \\
Sparsity & 0.695        & 0.920           & 0.788            & 0.406        & 0.838          & 0.537            & 0.313        & 0.715          & 0.412           \\ \midrule
        & \multicolumn{3}{c|}{Chongqing}                    & \multicolumn{3}{c|}{Chengdu}                     & \multicolumn{3}{c}{Xi'an}                        \\ \midrule
Shape   & (1488,31,51) & \multicolumn{2}{c|}{(4464,31,51)}  & (2928,9,9)   & (8784,9,9) & - & (2928,9,9)   & (8784,9,9) & -   \\
Mean    & 54.761       & 2.651           & 3.944            & 343.309      & 21.835         & -                & 227.624      & 12.897         & -               \\
Max     & 1572.0       & 495.0           & 411.8            & 3646.0       & 554.0          & -                & 2768.0       & 250.0          & -               \\
Sparsity & 0.411        & 0.727           & 0.477            & 0.028        & 0.102          & -                & 0.001        & 0.111          & - \\             
\bottomrule
\end{tabular}

\vspace{0.3cm}

\resizebox{11cm}{1.2cm}{
\begin{tabular}{l|llllll}
\toprule
Road Connectivity& Beijing & Shanghai & Shenzhen & Chongqing & Chengdu  & Xi'an    \\ \midrule
\# roads    & 180,488  & 151,124   & 69,612    & 31,138     & 6,846     & 11,810     \\
Max            & 65      & 67       & 32       & 27        & 35       & 23       \\
Mean ($\times$1e-4)    & 20.07  & 95.35   & 180.33  & 122.69   & 6752.02 & 4429.20 \\
Sparsity & 0.999  & 0.997   & 0.994   & 0.994    & 0.823 & 0.838 \\
\bottomrule
\end{tabular}}
\hspace{0.3cm}
\resizebox{6cm}{1.2cm}{\begin{tabular}{l|lll}
\toprule
Speed & Chengdu & Xi'an  & Hong Kong \\ \midrule
\# roads   & 1,676    & 792    & 618       \\
Max     & 170.6   & 149.22 & 111.0     \\
Mean    & 31.93   & 32.49  & 56.09    \\
Sparsity & 0.047   & 0.053  & 0.027     \\ 
\bottomrule
\end{tabular}}

\label{tab:stats}
\vspace{-0.4cm}
\end{table*}

\subsection{Taxi Flow Dataset}

By utilizing the taxi GPS points $\mathcal{G}_c$, we are able to derive inflow and outflow spatio-temporal tensors denoted as $\mathbf{x}^{in}_c$ and $\mathbf{x}^{out}_c$, respectively. These tensors are represented in $\mathbb{R}^{T_c\times W_c \times H_c}$  and capture the number of taxis entering and exiting each region at a given timestamp \cite{zhang2017deep}:
\begin{equation}
    \begin{aligned}
    \mathbf{x}^{in}_c(\tau, i, j) &= \sum_{tr\in \mathcal{G}_c} \sum_{(pos_k, t_k, occ_k)\in tr}\\
    &\mathbb{I}\left(t_k \in \tau \land pos_k\in r_c(i, j) \land pos_{k-1}\notin r_c(i, j)\right),\\
    \mathbf{x}^{out}_c(\tau,i, j) &= \sum_{tr\in \mathcal{G}_c}\sum_{(pos_k, t_k, occ_k)\in tr}\\
    &\mathbb{I}\left(t_{k+1} \in \tau \land pos_k\in r_c(i, j) \land pos_{k+1}\notin r_c(i, j)\right),
    \end{aligned}
\end{equation}
where $\mathbb{I}(X)$ is the indicator function. $\mathbb{I}(X) = 1$ iff $X$ is true. 

\subsection{Taxi Pickup and Idle Driving Dataset}
Efficient allocation of transportation resources often requires consideration of both the demand and supply data for taxis. Given taxi GPS points $\mathcal{G}_c$, we derive spatio-temporal tensors for the pickup and idle driving, denoted as $\mathbf{x}_c^{pickup}, \mathbf{x}_c^{idle}$, respectively. These tensors are instrumental in approximating the demands and supplies of taxis, thereby enabling us to allocate transportation resources more efficiently.
\begin{equation}
    \begin{aligned}
    \mathbf{x}_c^{pickup}(\tau, i, j) &= \sum_{tr\in\mathcal{G}_c}\sum_{(pos_k, t_k, occ_k)\in tr}\\
    &\mathbb{I}(t_k\in \tau \land occ_k = 1\land occ_{k-1} = 0),
    \end{aligned}
    \label{eqn:pickup}
\end{equation}
\begin{equation}
    \begin{aligned}
    \mathbf{x}_c^{idle}(\tau, i, j) &= \frac{1}{|\tau|}\sum_{tr\in\mathcal{G}_c}\sum_{(pos_k, t_k, occ_k)\in tr}\\
    &\mathbb{I}(t_k\in \tau \land occ_k = 0)(t_{k+1} - t_k),
    \end{aligned}
    \label{eqn:idle}
\end{equation}
where the length of a timestamp is denoted by $|\tau|$. Eqn. \eqref{eqn:pickup} computes the number of pickups of all taxis at the region $r_c(i, j)$ during the timestamp $\tau$, which serves as a lower bound for taxi demands. Meanwhile, Eqn. \eqref{eqn:idle} calculates the total fraction of time during the timestamp $\tau$ in which taxis are running in idle. For example, if a taxi $a$ was driving in idle in $r_c(i, j)$ for 10 minutes at timestamp $\tau$,  where $|\tau| = 30$ minutes, then $a$ contributes $1/3$ to $\mathbf{x}_c^{idle}(\tau, i, j)$. As taxis running in idle have the potential to serve passengers, Eqn. \eqref{eqn:idle} can be interpreted as an approximation of taxi supply. 

\subsection{POI Dataset}
 To obtain POI data for each city $c$, we utilized the AMap API\footnote{https://lbs.amap.com/} to gather  information in the form of $POI_c = \{(pos_{poi}, cat_{poi})\}$. Our analysis focused on 14 categories of POIs, which include \textit{scenic spots, medical and health, domestic services, residential area, finance, sports and leisure, culture and education, shopping, housing, governments and organizations, corporations, catering, transportation, and public services}. Each category was assigned an ID from 0 to 13. To construct a POI tensor $\mathbf{x}^{poi}_c\in \mathbb{R}^{W_c\times H_c\times 14}$ for each city $c$, we followed the procedure outlined below:
\begin{equation}
\resizebox{\columnwidth}{!}{
$\mathbf{x}^{poi}_c(i, j, cat) = \sum_{(pos_{poi}, cat_{poi})\in POI_c}\mathbb{I}(cat_{poi} = cat\land pos_{poi}\in r_c(i, j)).$
}
\end{equation}
$\mathbf{x}_c^{poi}(i, j, cat)$ is the number of POIs with category $cat$ in $r_c(i, j)$.

\subsection{Meteorology Dataset}
To collect weather data for each city, we obtained hourly records from the airport of each city through the rp5 website\footnote{https://rp5.ru/}. We then processed this data into a matrix format, denoted as $\mathbf{x}_c^{mtr}\in \mathbb{R}^{T_c/2\times 49}$. Each vector in this matrix is 49-dimensional and contains information on temperature, air pressure, humidity, visibility, and weather phenomena such as rain, thunderstorms, and fog. A detailed schema of the weather data can be found in the Appendix. 
\subsection{Road Connectivity Dataset}
To obtain road map information, we sourced data on highways and motorways from OpenStreetMaps\footnote{https://openstreetmap.org/}. Based on this data, we constructed region-wise adjacency matrices denoted as $\mathbf{A}^{conn}_c\in \mathbb{R}^{|\mathcal{R}_{c}|\times |\mathcal{R}_{c}|}$. These matrices were constructed as follows:
\begin{equation}
\mathbf{A}^{conn}_c(r_0, r_1) = \sum_{(pos_{0}, pos_{1})\in SEG_c}\mathbb{I}(pos_{0}\in r_0\land pos_{1}\in r_1),
\end{equation}
where $r_0, r_1\in \mathcal{R}_c$, the value of $\mathbf{A}^{conn}_{c}(r_0, r_1)$ represents the total number of roads connecting $r_0$ and $r_1$. As such, the adjacency matrix $\mathbf{A}_c^{conn}$ provides information on the transportation connectivity between different regions within city $c$.

\subsection{Traffic Speed Dataset}
 Real-time speed data was processed to obtain $\mathbf{x}^{speed}_c\in \mathbb{R}^{T_c\times |SEG_c|}$, where $T_c$ denotes the number of time steps and $|SEG_c|$ denotes the number of road segments within city $c$.
 
\begin{equation}
\mathbf{x}^{speed}_c(\tau, s) = \sum_{s\in SEG_c}\frac{\sum_{(t_{k},speed_{k})\in sp_{s}}\mathbb{I}(t_k \in \tau)\cdot speed_{k}}{\sum_{(t_k, speed_k)\in sp_s}\mathbb{I}(t_k \in \tau)}.
\end{equation}
Each element in $\mathbf{x}^{speed}_c$ represents the average traffic speed of a road segment $s$ at a particular timestamp $\tau$.

Table \ref{tab:stats} presents the statistics of the processed sub-datasets. It is worth noting that the cities in CityNet exhibit diverse properties. For instance, Beijing, Shanghai, Shenzhen, and Chongqing have large maps with more than 1000 regions ($|\mathcal{R}_c|>1000$), while Chengdu and Xi'an have small maps with less than 100 regions ($|\mathcal{R}_c|<100$). 

\section{Data mining and analysis}
\label{sec:mining}
In addition to the collection and processing of data, it is essential to identify and quantify the correlations between sub-datasets in CityNet to gain insights into the effective utilization of the multi-modal data. In this section, we leverage data mining tools to explore and visualize the relationships between service data and context data. By doing so, we aim to provide a deeper understanding of the interconnections between different data sources, which can inform the development of more effective transportation policies and strategies.

\begin{table*}[ht]
\centering
\caption{Adjusted Rand index (ARI) and adjusted mutual information (AMI) to evaluate the clustering assignments based on POIs and taxi mobility data. Both metrics range from -1 to 1, with larger values indicating a higher degree of coincidence between the two clusters.}
\vspace{-0.2cm}
\begin{tabular}{l|llll|llll|llll}
\toprule
    & Inflow & Outflow & Pickup & Idle-time & Inflow & Outflow & Pickup & Idle-time & Inflow & Outflow & Pickup & Idle-time \\ \cmidrule{2-13} 
    & \multicolumn{4}{c|}{Beijing}           & \multicolumn{4}{c|}{Shanghai}          & \multicolumn{4}{c}{Shenzhen}           \\ \midrule
ARI & 0.554  & 0.554   & 0.635   & 0.544     & 0.445  & 0.445   & 0.474   & 0.382     & 0.191  & 0.191   & 0.219   & 0.227     \\
AMI & 0.308  & 0.308   & 0.418   & 0.318     & 0.268  & 0.268   & 0.356   & 0.240     & 0.132  & 0.132   & 0.192   & 0.182     \\ \midrule
    & \multicolumn{4}{c|}{Chongqing}         & \multicolumn{4}{c|}{Chengdu}           & \multicolumn{4}{c}{Xi'an}               \\ \midrule
ARI & 0.471  & 0.471   & 0.624   & 0.541     & 0.197  & 0.196   & 0.167   & N/A          & 0.155  & 0.155   & 0.042   &  N/A         \\
AMI & 0.261  & 0.261   & 0.404   & 0.350     & 0.213  & 0.213   & 0.226   & N/A         & 0.164  & 0.164   & 0.05    & N/A   \\
\bottomrule
\end{tabular}

\label{tab:cluster_measure}
\end{table*}

\begin{figure*}[ht]
    \centering
    \subfloat[Beijing]{\includegraphics[width=\columnwidth,trim={2cm 0 2cm 0},clip]{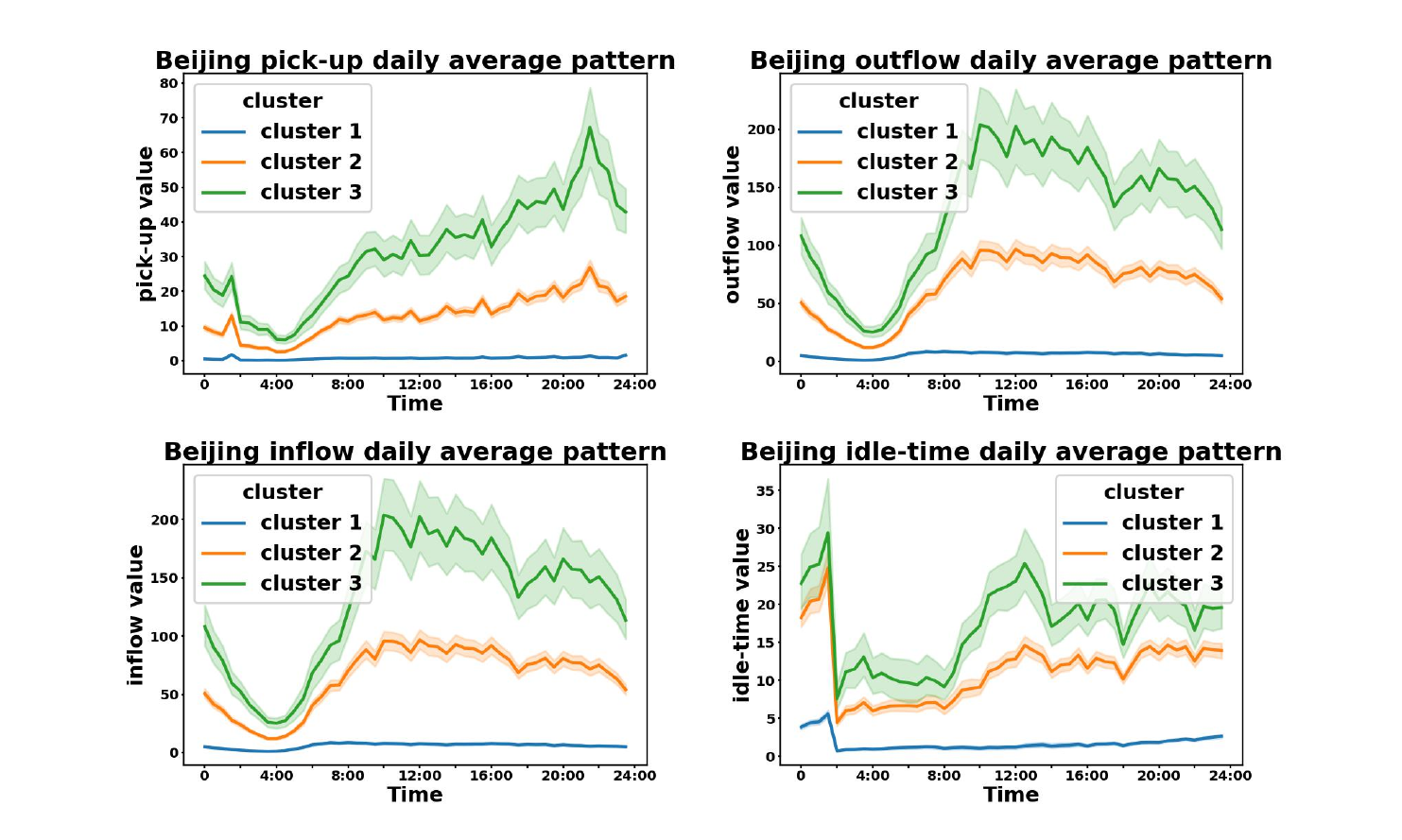}\label{fig:cluster-bj} }
    \hfill   
    \subfloat[Chengdu and Xi'an]{ \includegraphics[width=\columnwidth,trim={2cm 0 2cm 0},clip]{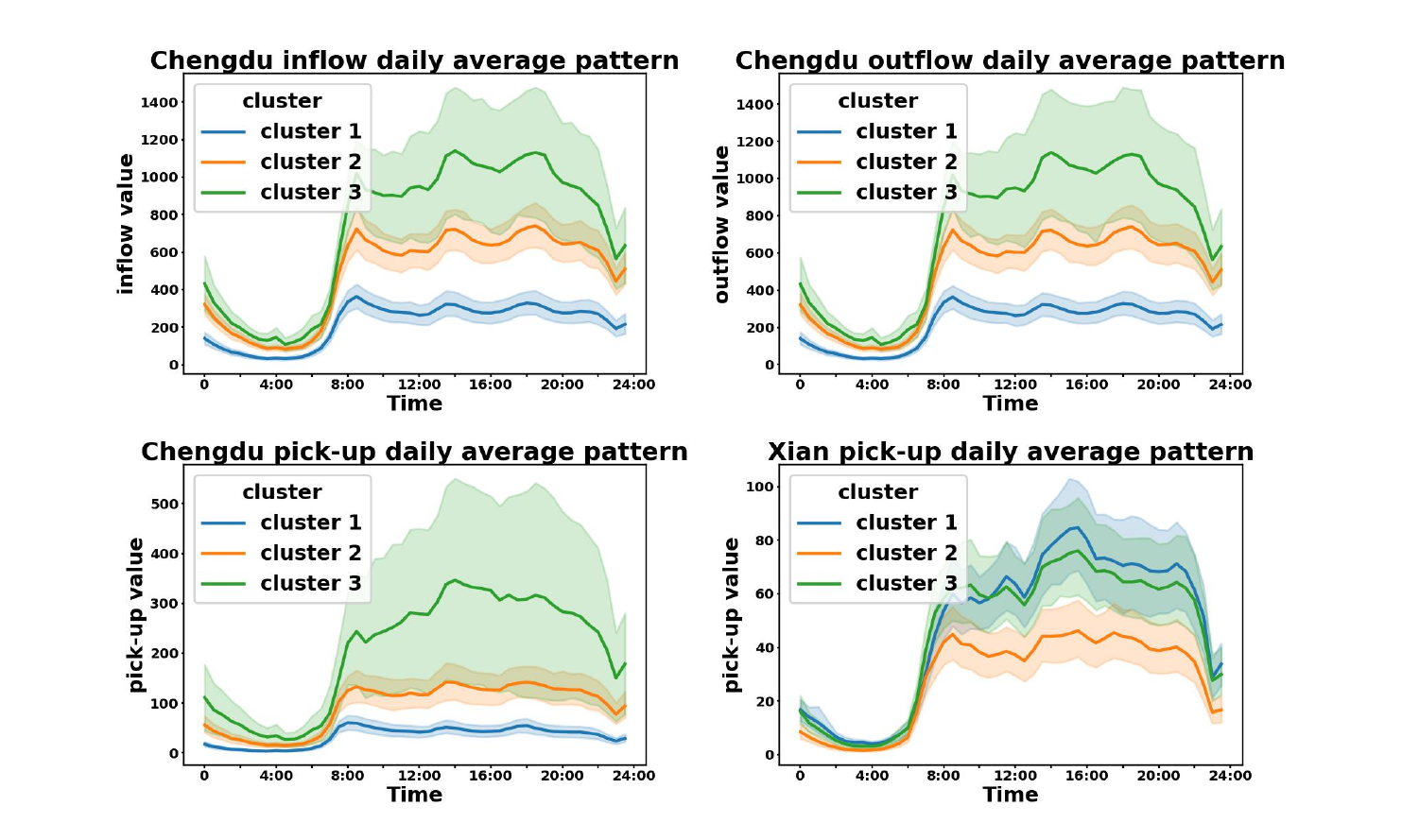} \label{fig:cluster-cdxa}}
    \caption{The average daily mobility pattern of taxi data for (a) Beijing, (b) Chengdu and Xi'an. To obtain these patterns, we aggregated all values at each timestamp from all days and represented the mean values at each timestamp using solid lines, while the standard deviations were represented using shades. \centering }
    \label{fig:taxi_poi}
\end{figure*}

\subsection{Correlation between POI and taxi mobility data}
\label{sec:poi_taxi}


 Taxi mobility data is expected to exhibit a strong correlation with Points of Interest (POIs). For instance, it is anticipated that the outflow of residential areas will experience a significant increase during morning peak hours, while taxi pickups should be more frequent in business areas during evening rush hours. To confirm this hypothesis, we implemented a clustering algorithm to cluster regions within city $c$ into $k=3$ disjoint clusters as below.
\begin{equation}
    \mathbb{C}_{c, poi}: \mathcal{R}_{c} = \mathcal{R}_{c, 1}\cup \mathcal{R}_{c, 2}\cup \cdots \cup \mathcal{R}_{c, k},
\end{equation}

We utilized the k-means algorithm to cluster regions within city $c$ based on their POI vectors ($\mathbf{x}_c^{poi}$). Additionally, we clustered regions based on the average daily pattern of taxi mobility data to obtain $\mathbb{C}_{c, in}$, $\mathbb{C}_{c, out}$, $\mathbb{C}_{c, pickup}$, $\mathbb{C}_{c, idle}$. To measure the similarity between the cluster assignments based on POIs ($\mathbb{C}_{c, poi}$) and those based on taxi mobility data ($\mathbb{C}_{c, in}$, etc.), we employed the adjusted Rand index (ARI) and the adjusted mutual information (AMI). The results, presented in Table \ref{tab:cluster_measure}, indicate positive correlations between the cluster assignments based on both POIs and taxi mobility data, thereby confirming the relationship between the two datasets.

The average regional daily patterns of taxi mobility data from each POI-based cluster in Beijing, Chengdu, and Xi'an are plotted in Fig. \ref{fig:taxi_poi}. As shown in Fig. 2\subref{fig:cluster-bj}, taxi mobility patterns in Beijing exhibit a high level of cohesion within each POI-based cluster, while remaining distinguishable across clusters. Conversely, Fig. 2\subref{fig:cluster-cdxa}, illustrates that clusters with higher inflow/outflow/pick-up values in Xi'an and Chengdu, two cities with relatively low ARI and AMI scores as reported in Table \ref{tab:cluster_measure}, demonstrate significant overlaps between adjacent clusters, which may be attributed to the limited number of regions in these cities. Nevertheless, Fig. 2\subref{fig:cluster-cdxa} still enables us to identify distinct clusters. 

Based on the aforementioned results, we draw the conclusion that regions with similar POI distributions also share similar taxi mobility patterns.

\subsection{Correlation between Meteorology and Traffic Speed Data}
\begin{table}[ht]
    \centering
    \caption{The regression coefficients between weather phenomena and traffic speed in Chengdu and Xi'an with a significance level of 1\% determined by a double-sided t-test.}
    \vspace{-0.2cm}
    \begin{tabular}{c|cc}
        \toprule
         & Xi'an & Chengdu \\
        \midrule
        Rain & -2.06 & -0.64\\
        Fog & -0.78 & -0.79 \\
        \bottomrule
    \end{tabular}
    \label{tab:speed-weather}
    \vspace{-0.4cm}
\end{table}


\begin{figure}[ht]
    \centering
    \subfloat[Rain]{\includegraphics[width=0.49\columnwidth]{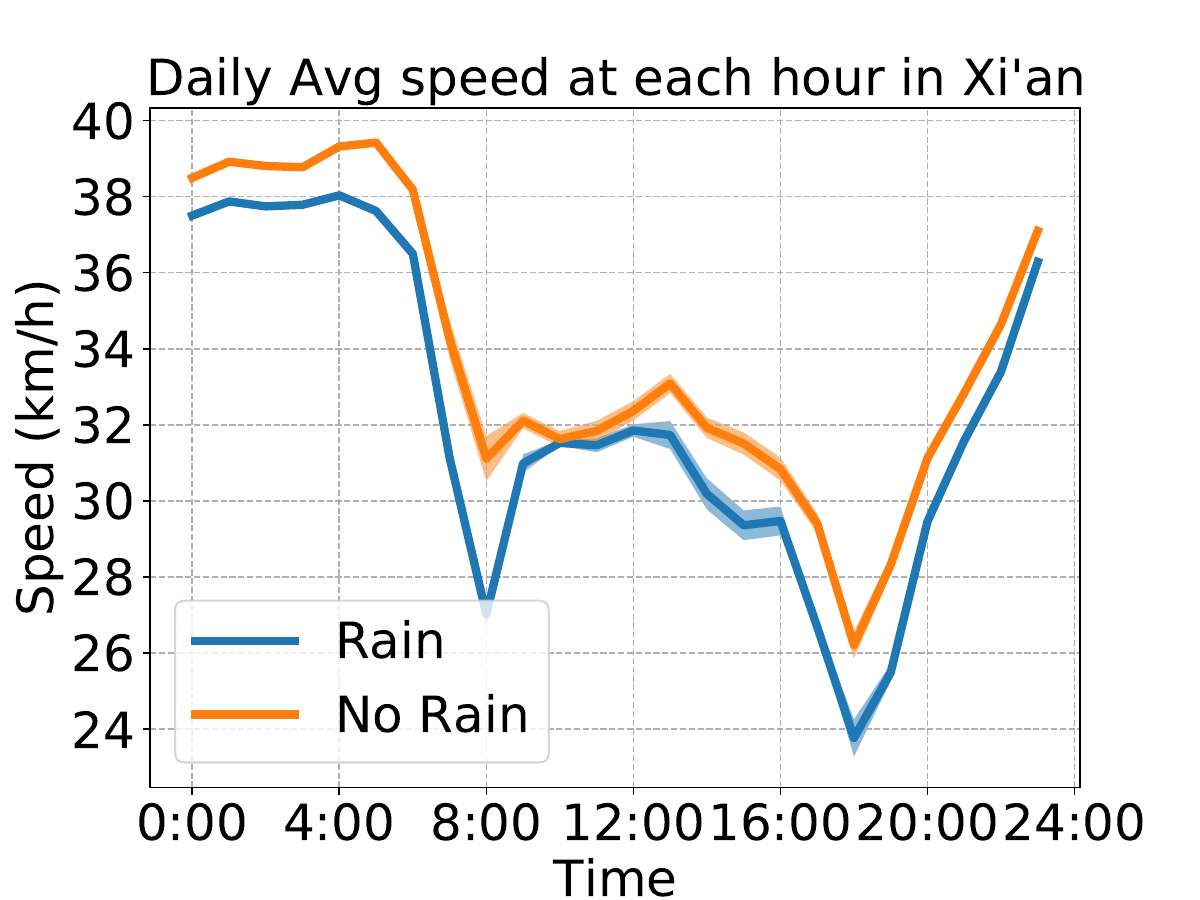}}
    \subfloat[Fog]{\includegraphics[width=0.49\columnwidth]{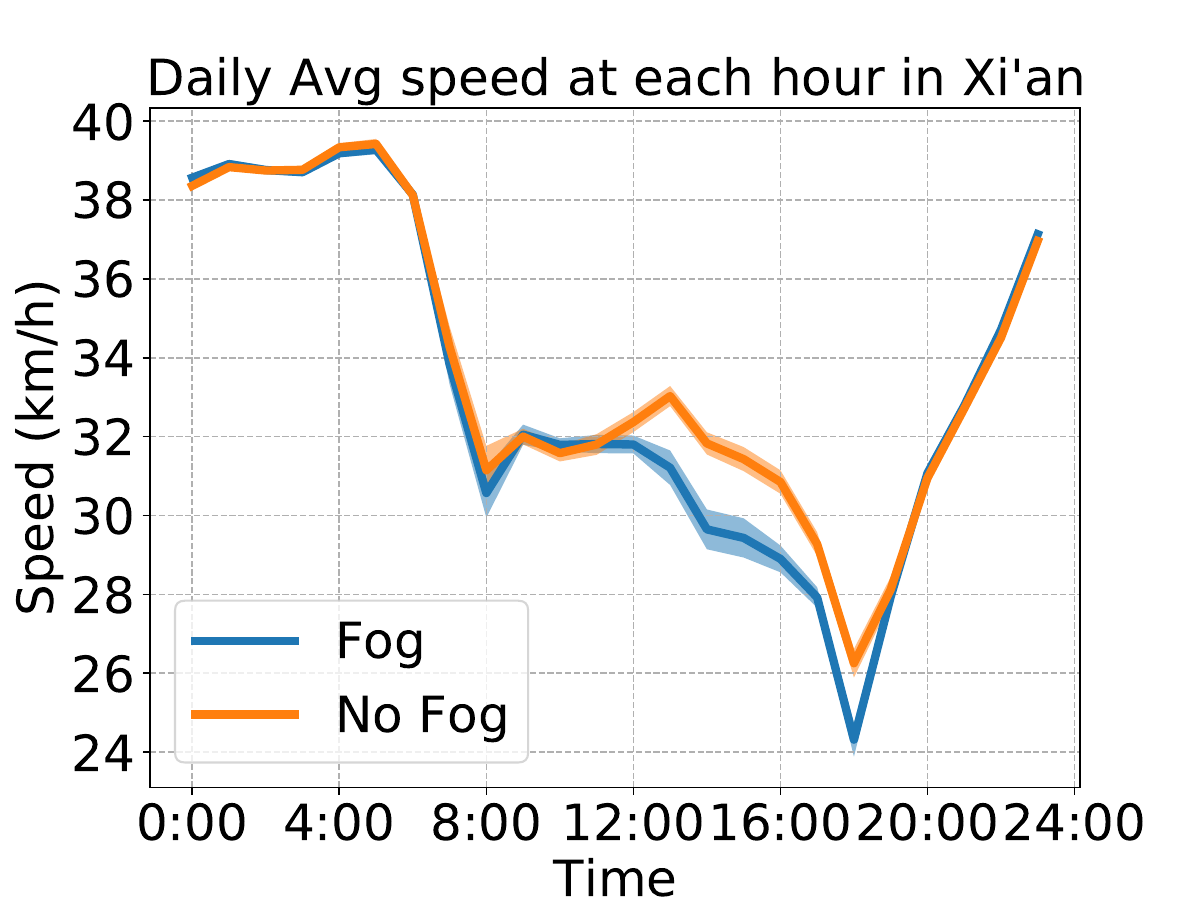}}
    \caption{The daily average speed in Xi'an with and without rain or fog and the shades in the figure represent half of the standard deviation. \centering }
    \label{fig:xian-speed-rain}
    \vspace{-0.2cm}
\end{figure}

The correlation between weather and traffic speed is widely acknowledged, as drivers tend to exercise caution and slow down in adverse weather conditions such as rain or fog. In order to empirically verify these correlations, we conducted a series of experimental studies.

Our research methodology encompasses both quantitative and qualitative studies. In our quantitative analysis, we utilized the least squares regression method to examine the relationship between weather phenomena and the average speed for each timestamp $\tau$, i.e. $\mathrm{MEAN}_s\left(\mathbf{x}_c^{speed}\left(\tau, s\right)\right)$ with the weather phenomena $\mathbf{x}^{mtr}_c(\tau)$. The regression coefficients are presented in Table \ref{tab:speed-weather}. Our findings indicate that adverse weather conditions such as rain and fog have a detrimental effect on the average traffic speed. To further illustrate this relationship, we plotted the daily average speeds in Xi'an with and without rain or fog in Fig. \ref{fig:xian-speed-rain}. The figure demonstrates consistently higher average speeds in the absence of rain, and higher average speeds during the day time when there was no fog.

\begin{table*}[ht]
\centering
\caption{The results for taxi mobility prediction of all models and tasks under 6 cities. The abbreviations ST-* and MT-* represent single-task and multi-task models, respectively. The lowest RMSE/MAE in each task is denoted in bold.}
\vspace{-0.2cm}
\resizebox{\linewidth}{!}{\begin{tabular}{ll|ll|llll|llll}
\toprule
City & Task & \multicolumn{2}{c|}{Non Deep Learning}      & \multicolumn{4}{c|}{Euclidean Deep Learning}     & \multicolumn{4}{c}{Graph Learning}    \\ 
\multicolumn{2}{c|}{RMSE/MAE} & ST-HA& ST-LR & ST-CNN & MT-CNN & ST-LSTM & MT-LSTM & ST-GCN & MT-GCN & ST-GAT & MT-GAT\\ 
\midrule
\multirow{4}[0]{*}{Beijing} & Inflow & 7.90/2.97 & 6.07/2.59 & 5.60/2.35 & 5.45/2.39 & 5.67/2.31 & 5.48/2.40 & 5.48/2.27 & \textbf{5.34/2.19} & 5.59/2.43 & 5.36/2.20 \\
& Outflow & 7.90/2.97 & 6.07/2.59 & 5.60/2.35 & 5.45/2.39 & 5.67/2.31 & 5.48/2.40 & 5.48/2.27 & \textbf{5.33/2.19} & 5.59/2.43 & 5.36/2.20\\
& Pickup  & 2.51/0.90 & 2.37/0.89 & 2.19/0.79 & 2.15/0.82 & 2.24/0.82 & 2.14/0.80 & 2.23/0.82 & 2.14/0.80 & 2.15/0.80 & \textbf{2.14/0.79}\\
& Idle & 0.73/0.31 & 0.61/0.25 & 0.57/0.24 & 0.54/0.25 & 0.60/0.25 & 0.52/0.22 & 0.55/0.24 & 0.51/0.23 & 0.55/0.24 & \textbf{0.50/0.22}\\
\midrule
\multirow{4}[0]{*}{Shanghai} & Inflow & 19.41/7.93 & 11.72/5.93 & \textbf{8.29/4.36} & 8.41/4.51 & 8.60/4.57 & 8.67/4.60 & 10.40/5.26 & 9.77/5.05 & 10.05/5.11 & 9.64/4.99 \\
& Outflow & 19.40/7.94 & 11.76/5.94 & 8.42/4.42 & \textbf{8.26/4.43} & 8.64/4.53 & 8.52/4.51 & 10.07/5.14 & 9.49/4.90 & 10.10/5.13 & 9.38/4.89 \\ 
& Pickup & 2.24/0.88 & 1.94/0.84 & 1.66/0.75 & 1.64/0.75 & 1.66/0.74 & \textbf{1.61/0.73} & 1.78/0.79 & 1.72/0.80 & 1.75/0.77 & 1.71/0.76 \\ 
& Idle & 1.23/0.54 & 0.89/0.41 & 0.86/0.44 & 0.75/0.41 & 0.85/0.41 & \textbf{0.73/0.36} & 0.86/0.42 & 0.79/0.38 & 0.81/0.40 & 0.78/0.38 \\
\midrule
\multirow{4}[0]{*}{Shenzhen} & Inflow & 43.79/18.97 & 24.06/12.23 & \textbf{17.62/9.41} & 18.28/9.75 & 17.94/9.57 & 17.75/9.65 & 22.18/11.61 & 20.84/10.91 & 22.35/11.46 & 20.52/10.87 \\
& Outflow & 43.85/18.97 & 24.08/12.23 & 18.08/9.54 & 18.45/9.53 & \textbf{17.70/9.54} & 17.80/9.64 & 22.08/11.33 & 20.82/10.81 & 21.99/11.28 & 20.59/10.82 \\ 
& Pickup & 6.78/2.48 & 4.91/2.09 & 4.29/1.83 & 4.24/1.76 & 4.00/1.75 & \textbf{3.99/1.76} & 4.39/2.02 & 4.24/1.86 & 4.50/1.92 & 4.21/1.84 \\ 
& Idle & 4.25/1.59 & 2.51/1.09 & 2.29/0.98 & 2.16/1.00 & \textbf{1.88/0.88} & 1.90/0.93 & 2.33/1.05 & 2.17/0.97 & 2.29/1.01 & 2.18/0.98 \\
\midrule
\multirow{4}[0]{*}{Chongqing} & Inflow & 37.40/14.97 & 21.19/9.91 & \textbf{12.65/6.48} & 12.67/6.52 & 14.39/7.29 & 13.84/7.08 & 18.54/6.59 & 17.12/8.46 & 16.53/8.49 & 15.98/7.93 \\
& Outflow & 37.36/14.87 & 21.23/9.85 & \textbf{12.90/6.60} & 12.99/6.60 & 14.30/7.10 & 13.68/6.95 & 18.75/8.87 & 16.91/8.29 & 18.12/9.12 & 15.94/7.93 \\ 
& Pickup & 7.64/3.03 & 6.34/2.69 & 5.25/2.24 & 5.15/2.10 & 5.21/2.18 & \textbf{5.02/2.10} & 5.77/2.45 & 5.49/2.33 & 5.72/2.45 & 5.34/2.28 \\ 
& Idle & 3.81/1.48 & 2.21/0.96 & \textbf{1.53/0.74} & 1.57/0.76 & 1.58/0.73 & 1.57/0.75 & 1.99/0.87 & 1.90/0.84 & 1.97/0.86 & 1.80/0.83 \\
\midrule
\multirow{3}[0]{*}{Xi'an} & Inflow & 87.00/55.28 & 43.69/29.72 & 27.79/19.20 & \textbf{25.77/17.85} & 28.13/19.39 & 26.33/17.94 & 34.77/23.86 & 31.70/21.77 & 31.62/21.67 & 30.67/21.59\\
& Outflow & 86.66/55.29 & 43.58/29.59 & 27.59/18.97 & \textbf{25.32/17.55} & 27.90/18.19 & 25.92/17.69 & 35.24/24.15 & 31.04/21.49 & 29.88/22.18 &28.96/20.12\\
& Pickup & 16.85/10.84 & 11.22/7.66 & 7.77/5.39 & \textbf{7.54/5.16} & 7.88/5.38 & 7.71/5.27 & 9.50/6.50 & 9.11/6.26 & 8.22/5.88 &8.05/5.26\\
\midrule
\multirow{3}[0]{*}{Chengdu} & Inflow & 117.5/71.1 & 58.92/37.72 & 31.22/21.09 & \textbf{30.02/20.40} & 34.50/23.57 & 32.78/22.42 & 45.10/29.49 & 41.27/27.62 & 40.02/27.27 & 38.47/25.84\\
& Outflow & 116.9/71.5 & 58.98/37.88 & 31.91/21.46 & \textbf{29.74/20.20} & 33.55/22.86 & 32.77/22.22 & 43.66/28.47 & 40.39/27.10 & 35.59/24.72 &35.20/24.58\\
& Pickup & 28.27/15.66 & 17.32/10.54 & 11.02/7.13 & \textbf{10.55/6.78} & 11.39/7.21 & 11.15/7.15 & 13.64/8.55 & 13.58/8.76 & 12.30/7.62 &11.24/7.38\\
\bottomrule
\end{tabular}}
\label{tab:taxi}
\vspace{-0.2cm}
\end{table*}

\begin{table}[ht]
\caption{Results for traffic speed prediction in Xi'an, Chengdu, and Hong Kong.}
\vspace{-0.2cm}
\begin{tabular}{l|lll}
\toprule
RMSE/MAE & Xi'an & Chengdu & Hong Kong \\ 
\midrule
HA       &8.7475/5.0312&7.9507/4.5202&6.784/3.8175\\
LR       &8.1432/4.5687&7.4246/4.1844&5.7752/3.3942\\ 
\midrule
GCN      &7.5872/4.1807&7.0598/3.9894&\textbf{5.5043/3.1405}\\
GAT      &\textbf{7.5217/4.0987}&\textbf{6.9622/3.8825}&5.5722/3.1465\\ 
\bottomrule
\end{tabular}
\label{tab:speed}
\end{table}

\section{Machine learning applications}
\label{sec:apps}

In this section, we present the empirical findings of machine learning tasks supported by CityNet, encompassing spatio-temporal predictions, transfer learning, and reinforcement learning. The primary objective of these experiments is to offer the following valuable insights: 
\begin{itemize}
    \item We present benchmark results for popular machine learning models applied to diverse sub-datasets in CityNet, showcasing their effectiveness in various prediction tasks.
    \item Our findings reveal a strong correlation among different prediction tasks concerning taxi mobility sub-datasets. By capitalizing on this mutual knowledge, we demonstrate that prediction accuracy for individual tasks can be enhanced through straightforward multi-task learning techniques such as weight sharing.
    \item we establish the presence of inter-city correlations that facilitate positive transfer learning, which can alleviate data scarcity.
    \item we demonstrate the efficacy of CityNet in supporting the task of taxi dispatching through reinforcement learning.
\end{itemize}

\subsection{Spatio-temporal Predictions}

 Spatio-temporal predictions of taxi mobility services are crucial tasks in the context of smart cities. Leveraging the data available in CityNet, we present benchmark results for two typical spatio-temporal prediction tasks.

\subsubsection{Taxi service prediction} 
\label{sec:taxi}
Our investigation focuses on the taxi service prediction task, encompassing both taxi flow \cite{zhang2017deep}, and demand-supply \cite{geng2019spatiotemporal} prediction\footnote{As previously stated in Sec. \ref{sec:architecture}, we utilize pickup and idle driving as proxies for demand and supply, respectively.}. 
\paragraph{Problem Formulation} Our approach to the taxi service prediction problem involves formulating it as a prediction function $f_c$, with the objective of achieving the lowest possible prediction error over future timestamps: 
\begin{equation}
\begin{aligned}
    \tilde{\mathbf{x}}_c^*(\tau) &= \tilde{f}_c\left(\left[\mathbf{x}_c^*(\tau-L),...,\mathbf{x}_c^*(\tau-1) \right] \right),\\
    f_c &= \arg\min_{\tilde{f}_c}\sum_{\tau\in \mathcal{T}_c}error\left(\tilde{\mathbf{x}}_c^*\left(\tau\right), \mathbf{x}_c^*\left(\tau\right)\right).\\
\end{aligned}
\label{eqn:stpred}
\end{equation}
where $\mathbf{x}_c^*$ refers to the spatio-temporal tensor to be predicted, which may include variables such as inflow, outflow, pickup, number of idle vehicles, etc. The error function, denoted as $error$, may take the form of RMSE or MAE, among others. We consider two distinct settings: the single-task and multi-task settings. In the single-task setting, $\mathbf{x}_c^*$ represents individual spatio-temporal tensors (e.g., in/outflow) while in the multi-task setting, it refers to stacked spatio-temporal tensors. Namely, 
\begin{equation}
    \mathbf{x}_c^{joint} = \mathrm{STACK}\left(\left[\mathbf{x}_c^{in}, \mathbf{x}_c^{out}, \mathbf{x}_c^{pickup}, \mathbf{x}_c^{idle}\right]\right)
\end{equation}
for the multi-task setting. To achieve multi-task learning for taxi service predictions, we employ direct weight sharing. We set the input length, denoted as $L$, to 5, which corresponds to a duration of 2.5 hours. \footnote{It is worth noting that for the pickup and idle driving datasets, we aggregate 10-minute timestamps into 30-minute timestamps and use the resulting aggregated tensors for prediction.}

\paragraph{Methods}
Our investigation centers on the following methods for spatio-temporal predictions of taxi mobility.
\begin{itemize}
    \item \textbf{HA} (History Average): Mean of historical values of the data.
    \item \textbf{LR} (Linear Regression): We employ linear regression on historical values of the data, utilizing Ridge with a regularization parameter of 0.01. 
    \item \textbf{CNN} (Convolutional Neural Network): Our CNN is based on STResNet \cite{zhang2017deep} and implemented using 6 residual blocks with 64 $3\times 3$ filters. We incorporate batch normalization into the architecture. 
    \item \textbf{LSTM}: Our LSTM model is based on ST-net \cite{yao2019learning}, which involves applying convolutions to the input tensors before using LSTM to capture temporal dependencies. We employ 3 residual blocks with batch normalization for the convolutions, while for the LSTM component, we use a single-layer LSTM with 256 hidden units. 
    \item \textbf{GCN} (Graph Convolutional Network) and \textbf{GAT} (Graph Attention Network): We apply GCN \cite{kipf2016semi} and GAT \cite{velivckovic2017graph}, with the input features being either the historical values (single-task) or stacked historical values of all tasks (multi-task) of the input data. For GCN, we denote the adjacency matrix as $\mathbf{A} = \left(\mathbf{A}_c^{conn} + \mathbf{A}_c^{poi} + \mathbf{A}_c^{neigh}\right)/3$ following \cite{geng2019spatiotemporal}, while for GAT, we use $\hat{\mathbf{A}} = \mathbf{D}^{-1/2}\mathbf{A}\mathbf{D}^{-1/2}$ as the normalized adjacency matrix. 
\end{itemize}
Among the methods studied, HA and LR are traditional time series forecasting models, while CNN and LSTM are deep learning models designed for Euclidean structures (such as grid networks). By contrast, GCN and GAT are graph deep learning models that leverage additional region-wise connections, such as POI and road connectivity.

\paragraph{Experimental Settings}
 To evaluate the performance of the models, we split the data into training (70\%)  validation (15\%), and testing (15\%) sets. All data are normalized to the range of $[0, 1]$ using min-max normalization and will be recovered for evaluation. The metrics employed for evaluation are RMSE and MAE following \cite{yu2018spatio}. . 

Table \ref{tab:taxi} illustrates the comprehensive performance of all the models examined in both the single-task (ST-*) and the multi-task (MT-*) settings. Our analysis of the results leads us to the following observations:
\begin{itemize}
    \item \textbf{Deep Learning or Not}: When provided with ample data (e.g., 10 days, 1-2 months), deep learning models such as CNN, LSTM, GCN, and GAT exhibit superior performance compared to traditional time-series forecasting methods such as HA and LR. This highlights the potency of deep learning in spatio-temporal predictions and the benefits of utilizing information in both Euclidean and non-Euclidean spaces.
    \item \textbf{Multi-task or Not}: Out of the 22 tasks examined, multi-task models exhibit the lowest RMSE in \textbf{15 (68.2\%)} tasks and the lowest MAE in \textbf{19 (86.4\%)} tasks. Our findings suggest that a simple multi-task learning approach, utilizing weight sharing, can enhance taxi service predictions by establishing connections among diverse taxi mobility data. This study paves the way for future research in multi-task spatio-temporal predictions. 
    \item \textbf{Graph Models or Not}: Among the 16 tasks conducted in Beijing, Shanghai, Shenzhen, Chongqing, GNN models exhibit the lowest RMSE in \textbf{8 (50\%)} tasks and the lowest MAE in \textbf{all} tasks except for Xi'an and Chengdu where CNN models outperform all other models in all tasks. Our analysis, as presented in Table \ref{tab:stats}, reveals that Xi'an and Chengdu have significantly smaller map sizes than the other cities. We conclude that in cities with large map sizes, incorporating graphs to model region-wise connections can significantly enhance prediction accuracy, while in cities with small map sizes, incorporating graphs may actually impair performance. One possible explanation for this phenomenon is that graphs built on a limited number of regions tend to be dense, resulting in over-smoothing \cite{li2018deeper}. 
\end{itemize}

\subsubsection{Traffic speed prediction}
Next, we investigate the problem of predicting traffic speed \cite{yu2018spatio,li2017diffusion}. We formulate this problem as learning a \textbf{multi-step} prediction function $f_c(\cdot)$:
\begin{gather}
\begin{aligned}
    &\left[\tilde{\mathbf{x}}_c^s(\tau),\,...\,,\tilde{\mathbf{x}}_c^s(\tau+L_{o}-1)\right]  \\ &= \tilde{f}_c\left(\left[\mathbf{x}_c^s(\tau-L_{i}),...,\mathbf{x}_c^s(\tau-1) \right] \right),\\
    & f_c = \arg\min_{\tilde{f}_c}\sum_{\tau\in \mathcal{T}_c} error \left(\tilde{\mathbf{x}}_c^s\left(\tau\right), \mathbf{x}_c^s\left(\tau\right)\right),
\end{aligned}
\label{eqn:trafpred}
\end{gather}
where $\mathbf{x}_{c}^s$ is an abbreviation for the traffic speed sub-dataset $\mathbf{x}^{speed}_c$. For this experiment, we set the input length  $L_{i}$ to 12 (equivalent to 2 hours) and the output length $L_{o}$ to 6 (equivalent to 1 hour).

\begin{table*}[ht]
\caption{The results of inter-city transfer learning from source domains (Beijing, Shanghai, and Xi'an) to target domains (Shenzhen, Chongqing, and Chengdu). The lowest RMSE/MAE using limited target data is highlighted in bold. The results under full data and 3-day data represent the lower and upper bounds for the errors of the corresponding tasks.}
\label{tab:transfer}
\vspace{-0.2cm}
\resizebox{\linewidth}{!}{\begin{tabular}{ll|ll|ll|lll|lll}
\toprule
\multirow{2}{*}{Target City} & \multirow{2}{*}{Task} & \multicolumn{2}{c|}{Full Data (Lower bound)} & \multicolumn{2}{c|}{3-day Data (Upper bound)} & \multicolumn{3}{c|}{Fine-tuning/Source City} & \multicolumn{3}{c}{RegionTrans/Source City}\\ 
& & LR & LSTM & LR & LSTM & Beijing & Shanghai & Xi'an & Beijing & Shanghai & Xi'an\\ 
\midrule
\multirow{4}[0]{*}{Shenzhen} & Inflow & 24.06/12.23 & 17.94/9.57 & 24.07/12.27 & 22.63/12.83 & 19.78/10.58 & 20.90/11.09 & N/A & \textbf{19.67/10.40} & 20.59/10.93 & N/A \\
& Outflow & 24.08/12.23 & 17.80/9.64 & 24.09/12.27 & 22.75/12.87 & 19.82/10.58 & 20.95/11.09 & N/A & \textbf{19.62/10.49} & 20.70/10.95 & N/A \\ 
& Pickup & 4.91/2.09 & 3.99/1.76 & 4.92/2.09 & 4.52/2.07 & 4.25/1.86 & 4.37/1.92 & N/A & \textbf{4.21/1.82} & 4.31/1.89 & N/A \\ 
& Idle & 2.51/1.09 & 1.90/0.94 & 2.51/1.09 & 2.39/1.17 & \textbf{2.12/0.97} & 2.16/1.10 & N/A & 2.12/0.97 & 2.13/1.00 & N/A \\
\midrule
\multirow{4}[0]{*}{Chongqing} & Inflow & 21.19/9.91 & 13.84/7.08 & 21.20/9.95 & 18.45/9.81 & 16.10/7.98 & 16.74/8.23 & N/A & \textbf{15.87/7.90} & 16.59/8.15 & N/A \\
& Outflow & 21.24/9.85 & 13.68/6.95 & 21.24/9.86 & 18.22/9.69 & 15.85/7.83 & 16.61/8.13 & N/A & \textbf{15.72/7.83} & 16.46/8.08 & N/A \\ 
& Pickup & 6.34/2.69 & 5.02/2.10 & 6.34/2.69 & 5.64/2.52 & 5.42/2.23 & 5.45/2.30 & N/A & \textbf{5.37/2.20} & 5.42/2.30 & N/A \\ 
& Idle & 2.21/0.96 & 1.57/0.75 & 2.21/0.96 & 2.03/0.98 & \textbf{1.75/0.81} & 1.80/0.83 & N/A & 1.77/0.82 & 1.77/0.81 & N/A \\
\midrule
\multirow{3}[0]{*}{Chengdu} & Inflow & 58.92/37.72 & 32.79/22.42 & 59.29/38.45 & 48.39/33.12 & 43.49/28.88 & 46.30/30.08 & 45.24/29.69 & \textbf{43.43/28.60} & 45.74/29.92 & 44.43/29.51\\
& Outflow & 58.98/37.88 & 33.55/22.86 & 59.37/38.61 & 47.77/32.69 & 43.12/\textbf{28.44} & 46.06/30.12 & 44.66/29.35 & \textbf{42.73}/28.66 & 45.59/29.86 & 43.69/29.17\\
& Pickup & 17.32/10.54 & 11.15/7.15 & 17.40/10.69 & 15.21/10.08 & 13.78/8.79 & 14.39/9.12  & 14.07/9.01 & \textbf{13.51/8.48} & 14.21/9.03 & 13.89/9.06\\
\bottomrule
\end{tabular}}

\end{table*}
In this study, we examine four methods: HA, LR, GCN, and GAT, as described in Section \ref{sec:apps}. For $i,j\in SEG_{c}$, where $i=(pos_{0}^{i},pos_{1}^{i})$ and $j = (pos_{0}^{j},pos_{1}^{j})$, the adjacency matrix $\mathbf{A}$ for GNN models is calculated as 
\begin{equation}
\mathbf{A}(i,j)=
\left\{
             \begin{array}{lr}
             1, & pos_{0}^{i}=pos_{1}^{j}\text{ or }pos_{1}^{i}=pos_{0}^{j} \\
             0, & \text{otherwise}
             \end{array}.
\right.
\end{equation}
Table \ref{tab:speed} presents the comprehensive performance evaluation of all the examined models for the traffic speed prediction tasks across three cities.

\subsection{Inter-city Transfer Learning}
As depicted in Table \ref{tab:taxi}, deep learning models can generate highly accurate predictions when provided with ample data. However, the level of digitization varies significantly among cities, and it is likely that many cities may not be able to construct accurate deep learning prediction models due to a lack of data. One effective solution to this problem is transfer learning \cite{pan2009survey}, which leverages knowledge from a source domain with abundant data to a target domain with limited data. In our case, this involves transferring knowledge from one city to another. Therefore, we conduct transfer learning experiments on CityNet to demonstrate that inter-city connections can facilitate positive knowledge transfer and to establish benchmarks for future research on inter-city transfer learning.

In this study, we examine the problem of inter-city transfer learning on taxi mobility sub-datasets. We formulate this problem as learning a function $f_{c_s, c_t}$ that that minimizes the prediction error on the target city $c_t$, with the assistance of data from the source city $c_s$: 
\begin{equation}
\begin{aligned}
    \tilde{\mathbf{x}}_{c_t}^*(\tau) &= \tilde{f}_{c_s, c_t}\left(\left[\mathbf{x}_{c_t}^*(\tau-L),...,\mathbf{x}_{c_t}^*(\tau-1) \right], \mathbf{x}_{c_s}^* \right),\\
    f_{c_s, c_t} &= \arg\min_{\tilde{f}_{c_s, c_t}}\sum_{\tau\in \mathcal{T}_{c_t}}error\left(\tilde{\mathbf{x}}_{c_t}^*(\tau), \mathbf{x}_{c_t}^*(\tau)\right).\\
    &|\mathbf{x}_{c_t}^*|\ll |\mathbf{x}_{c_s}^*|,
\end{aligned}
\end{equation}
with the notations following Eqn. \eqref{eqn:stpred}.

\paragraph{Experimental Settings} To address this problem, we utilize LSTM as the base model, which is similar to ST-net in MetaST \cite{yao2019learning}, and adopt a multi-task learning approach. We select Beijing and Shanghai as the source cities for transfer learning tasks in cities with large map sizes, and Xi'an as the source city for the transfer learning tasks in cities with small map sizes, with the remaining cities in CityNet serving as target cities. For all target cities, we train the model using a three-day interval (Thursday-Friday-Saturday) and maintain the same validation and test data. 

\paragraph{Methods} We examine the following transfer learning methods: 
\begin{itemize}
    \item \textbf{Fine-tuning}: Our approach involves training a model on the source city and subsequently fine-tuning it using a limited amount of target data. 
    \item \textbf{RegionTrans}: The RegionTrans algorithm \cite{wang2019cross} utilizes time-series (S-match) or auxiliary data\footnote{In this paper, context data serve as auxiliary data.} (A-match) correlation to compute a matching between the source region $\mathcal{R}_{c_s}$ and target regions $\mathcal{R}_{c_t}$. This matching is then employed as a regularizer, ensuring that similar regions share similar features. In our experimental setup, we leverage POI vectors to compute the matching, which is part of the A-match approach.
\end{itemize}
\paragraph{Results} Table \ref{tab:transfer} presents the results of our inter-city transfer learning experiments. Specifically, we report the results obtained by training our models using both full and 3-day target data, which correspond to the lower and upper bounds of errors, respectively. Furthermore, we also include the results of fine-tuning and RegionTrans methods. Based on the results, we obtain the following observations: 
\begin{itemize}
    \item \textbf{Degradation under data scarcity.} Our findings reveal that when only 3-day training data are available, non-deep learning models such as LR achieve similar performances as compared to using full data, whereas LSTM models suffer from an increased error rate of 50\%, as observed in the case of Chengdu. This suggests that deep learning models exhibit greater sensitivity to the amount of training data as compared to non-deep learning models.
    \item \textbf{Inter-city correlations.} Our results demonstrate that transfer learning leads to error reductions in all source-target pairs, as compared to using target data only. Notably, the largest reduction of approximately 15\% is observed in the case of Shenzhen and Chongqing. These findings suggest that there exist sufficient correlations between cities in CityNet, such that knowledge learned for one city can be effectively transferred and positively impact the performance in another city. 
    \item \textbf{Impact of Context Data.} By leveraging POI data for region matching, our proposed RegionTrans method achieves lower error rates than fine-tuning in most cases. This finding, coupled with the results presented in Section \ref{sec:poi_taxi}, underscores the importance of multi-modal data in CityNet and verifies the connection between context and service data. 
    \item \textbf{Domain Selection.} Our experimental results consistently demonstrate that using Beijing as the source city yields the best performance, irrespective of the target city and the choice of algorithms. One possible explanation for this observation is that Beijing comprises the highest number of regions, and therefore exhibits a more complex regional service pattern as compared to other cities. 
\end{itemize}

\subsection{Reinforcement Learning}

 Efficient taxi allocation is crucial for the passenger transportation services in smart cities. To address this challenge, we leverage the data available in CityNet and present benchmarks for the taxi dispatching task. In this task, operators are responsible for dispatching available taxis to waiting passengers in real-time with the objective of maximizing the long-term total revenue of the taxi system.

\textit{Problem Statement}. To address the taxi dispatching task, we learn a real-time dispatching policy based on historical passenger requests. At every timestamp $\tau$,  we use this policy to dispatch available taxis to current passengers, with the aim of maximizing the total revenue of all taxis in the long run. To achieve this, we divide the city into uniform hexagonal grids, as opposed to square grids used in previous studies \cite{tang2019reinforcement,xu2018reinforcement}.

\textit{Methods}. We investigate the following methods for taxi dispatching.

\begin{itemize}
    \item \textbf{Greedy algorithm} is a heuristic approach that dispatches the closest available taxi to each passenger, one by one.
    
    \item \textbf{LLD algorithm } is an optimization-based approach formulated by Eqn. \eqref{eqn:dispatching}, where $a_{ij} = 1$ if taxi $j$ is dispatched to passenger $i$ and 0 otherwise; Here, $A(i,j)$ represents the immediate revenue earned by taxi $j$after transporting passenger $i$ to their destination. The definition of immediate revenue follows the approach presented in \cite{xu2018reinforcement}.
\begin{equation}
\begin{split}
& \max_{a_{ij}} \sum_{i=0}^{m}\sum_{j=0}^{n} A(i, j)a_{ij}\\
& s.t. \left\{\begin{matrix}
\sum_{i=0}^{m} a_{ij}=1, j=1, 2, 3, ..., n \\
\\ 
\sum_{j=0}^{n} a_{ij}=1, i=1, 2, 3, ..., m
\end{matrix}\right.
\end{split}
\label{eqn:dispatching}
\end{equation}
    
\item \textbf{LPA algorithm} is a reinforcement learning-based approach \cite{xu2018reinforcement}. We first adopt SARSA \cite{xu2018reinforcement} to learn the expected long-term revenue of each grid in each period. Based on these expected revenues, we dispatch taxis to passengers using the same optimization formulation as Eqn. \eqref{eqn:dispatching}, with the exception that we replace $A(i, j)$ with the scores learned by SARSA. Unlike other methods that focus on immediate revenues in the current execution, LPA aims to maximize the total revenue of the system in the long run.

\end{itemize}

\paragraph{Experimental Settings}  To evaluate the performance of these methods, we implement a simulator based on historical request data using the approach presented in \cite{xu2018reinforcement,zhou2019multi}. This simulator allows us to simulate the operation of the taxi system and assess the effectiveness of each method.

Table \ref{tab:reinforcement} presents the taxi dispatching results for Chengdu, where the completion rate denotes the ratio of completed requests within all requests, and accumulated revenue represents the total revenue earned by all taxis throughout the day. Based on the experimental results, we draw the following conclusions:

\begin{itemize}
    \item The Greedy algorithm, which does not consider any global optimization targets, performs the worst compared to LLD and LPA. Taking global optimization targets into consideration leads to a significant improvement in performance, with completion rates improving by 5\%$\sim$20\% and revenue increasing by 2\%$\sim$8\%.This highlights the importance of incorporating global optimization targets in the dispatching process.
    \item  Our experimental results demonstrate that LPA outperforms LLD in most cases. This can be attributed to the fact that LPA optimizes the expected long-term revenues at each dispatching round, while LLD only focuses on the immediate reward. As a result, LPA is better suited for maximizing the total revenue of the system in the long run, and is expected to compare favorably against LLD. 
\end{itemize}

\begin{table}[ht]
\caption{Results for taxi dispatching in Chengdu}
\label{tab:reinforcement}
\vspace{-0.2cm}
\resizebox{\columnwidth}{!}{
\begin{tabular}{c|ccc|ccc}
\toprule
\multirow{2}[0]{*}{Date} & \multicolumn{3}{c|}{Completion rate} & \multicolumn{3}{c}{Accumulated revenue} \\

\cmidrule{2-7}

 & Greedy & LLD & LPA & Greedy & LLD & LPA \\ 
\midrule
Oct. 1st & 0.689 & \textbf{0.734} & 0.718 & 1,017,144 & \textbf{1,035,724} & 1,016,410 \\
Oct. 2nd & 0.705 & \textbf{0.765} & 0.752 & 969,340 & \textbf{984,424} & 975,157 \\ 

Oct. 3rd & 0.680 & 0.734 & \textbf{0.747} & 943,422 & 961,263 & \textbf{969,341} \\
Oct. 4th & 0.695 & 0.810 & \textbf{0.830} & 965,896 & 1,034,539 & \textbf{1,041,978} \\
Oct. 5th & 0.708 & 0.782 & \textbf{0.811} & 984,225 & 1,021,727 & \textbf{1,032,165} \\
Oct. 6th & 0.735 & \textbf{0.832} & 0.825 & 979,356 & \textbf{1,028,292} & 1,020,174 \\
Oct. 7th & 0.774 & 0.872 & \textbf{0.889} & 932,641 & 977,996 & \textbf{980,397} \\
\bottomrule
\end{tabular}}

\end{table}

\section{Conclusion and opportunities}
\label{sec:conclusion}
In the present study, we have introduced CityNet, a multi-modal dataset specifically designed for urban computing in smart cities, which incorporates spatio-temporally aligned urban data from multiple cities and diverse tasks. To the best of our knowledge, CityNet is the first dataset of its kind, which provides a comprehensive and integrated view of urban data from various sources. Through the use of data mining and visualization tools, we have demonstrated the significance of multi-modal urban data and have highlighted the connections between service and context data. Furthermore, we have presented extensive experimental results on spatio-temporal predictions, transfer learning, and reinforcement learning, which demonstrate the potential of CityNet as a versatile benchmark for various research topics.

CityNet provides ample opportunities for future research, including but not limited to:
\begin{itemize}
    \item \textbf{Transfer learning}: Firstly, it can serve as an ideal testbed for transfer learning algorithms, including meta-learning \cite{yao2019learning}, AutoML \cite{li2020autost}, and transfer learning on spatio-temporal graphs under homogeneous or heterogeneous representations. In the field of urban computing, it is highly probable that the knowledge required for different tasks, cities, or time intervals is correlated. By leveraging this transferable knowledge across domains with this multi-city, multi-task data, CityNet can help researcher alleviate the data scarcity problems that arise in newly-built or under-developed cities.

    
    \item \textbf{Federated learning}: Secondly, CityNet is an appropriate dataset to investigate various federated learning topics under different settings, with each party holding data from one source or one city. Urban data is usually generated by a multitude of human activities and stored by diverse  stakeholders, such as organizations, companies, and the government. However, due to data privacy regulations or the need to protect commercial interests, collaborations between these stakeholders should be conducted in a privacy-preserving manner. Federated learning (FL) \cite{yang2019federated} could provide an effective solution for enabling privacy-preserving multi-party collaboration and can be specifically investigated using CityNet, with its data from multiple cities and sources.
    
    \item \textbf{Managerial insight}: Thirdly, we hope that CityNet can be used for applications providing explicit insight into urban management, including causal inference \cite{liu2011discovering}, uncertainty estimations, and explainable machine learning. By mining and comprehending the social effects derived from urban data, city governors can make informed decisions on urban management, leading to wiser and more effective strategies.
\end{itemize}

We anticipate that CityNet will serve as a catalyst for further research efforts in the field of machine learning for urban computing. In our future work, we aim to expand CityNet by incorporating additional sources of data, such as subways and buses, thereby enhancing its comprehensiveness and usefulness.

\bibliographystyle{IEEEtran}
\bibliography{ref}

\begin{thebibliography}{10}
\providecommand{\url}[1]{#1}
\csname url@samestyle\endcsname
\providecommand{\newblock}{\relax}
\providecommand{\bibinfo}[2]{#2}
\providecommand{\BIBentrySTDinterwordspacing}{\spaceskip=0pt\relax}
\providecommand{\BIBentryALTinterwordstretchfactor}{4}
\providecommand{\BIBentryALTinterwordspacing}{\spaceskip=\fontdimen2\font plus
\BIBentryALTinterwordstretchfactor\fontdimen3\font minus \fontdimen4\font\relax}
\providecommand{\BIBforeignlanguage}[2]{{%
\expandafter\ifx\csname l@#1\endcsname\relax
\typeout{** WARNING: IEEEtran.bst: No hyphenation pattern has been}%
\typeout{** loaded for the language `#1'. Using the pattern for}%
\typeout{** the default language instead.}%
\else
\language=\csname l@#1\endcsname
\fi
#2}}
\providecommand{\BIBdecl}{\relax}
\BIBdecl

\bibitem{zheng2014urban}
Y.~Zheng, L.~Capra, O.~Wolfson, and H.~Yang, ``Urban computing: concepts, methodologies, and applications,'' \emph{ACM Transactions on Intelligent Systems and Technology (TIST)}, vol.~5, no.~3, pp. 1--55, 2014.

\bibitem{zhang2016dnn}
J.~Zhang, Y.~Zheng, D.~Qi, R.~Li, and X.~Yi, ``Dnn-based prediction model for spatio-temporal data,'' in \emph{Proceedings of the 24th ACM SIGSPATIAL International Conference on Advances in Geographic Information Systems}, 2016, pp. 1--4.

\bibitem{li2017diffusion}
Y.~Li, R.~Yu, C.~Shahabi, and Y.~Liu, ``Diffusion convolutional recurrent neural network: Data-driven traffic forecasting,'' \emph{arXiv preprint arXiv:1707.01926}, 2017.

\bibitem{wang2019cross}
L.~Wang, X.~Geng, X.~Ma, F.~Liu, and Q.~Yang, ``Cross-city transfer learning for deep spatio-temporal prediction,'' in \emph{Proceedings of the 28th International Joint Conference on Artificial Intelligence}.\hskip 1em plus 0.5em minus 0.4em\relax AAAI Press, 2019, pp. 1893--1899.

\bibitem{yao2019learning}
H.~Yao, Y.~Liu, Y.~Wei, X.~Tang, and Z.~Li, ``Learning from multiple cities: A meta-learning approach for spatial-temporal prediction,'' in \emph{The World Wide Web Conference}, 2019, pp. 2181--2191.

\bibitem{xu2018reinforcement}
Z.~Xu, Z.~Li, Q.~Guan, D.~Zhang, Q.~Li, J.~Nan, C.~Liu, W.~Bian, and J.~Ye, ``Large-scale order dispatch in on-demand ride-hailing platforms: A learning and planning approach,'' in \emph{Proceedings of the 24th ACM SIGKDD International Conference on Knowledge Discovery and Data Mining}, 2018, pp. 905--913.

\bibitem{li2019efficient}
M.~Li, Z.~Qin, Y.~Jiao, Y.~Yang, J.~Wang, C.~Wang, G.~Wu, and J.~Ye, ``Efficient ridesharing order dispatching with mean field multi-agent reinforcement learning,'' in \emph{The World Wide Web Conference}, 2019, pp. 983--994.

\bibitem{chen2001freeway}
C.~Chen, K.~Petty, A.~Skabardonis, P.~Varaiya, and Z.~Jia, ``Freeway performance measurement system: mining loop detector data,'' \emph{Transportation research record}, vol. 1748, no.~1, pp. 96--102, 2001.

\bibitem{H2014Big}
H.~Jagadish, J.~Gehrke, A.~Labrinidis, Y.~Papakonstantinou, J.~M. Patel, R.~Ramakrishnan, and C.~Shahabi, ``Big data and its technical challenges,'' \emph{Communications of the Acm}, vol.~57, no.~7, pp. 86--94, 2014.

\bibitem{ferreira2013visual}
N.~Ferreira, J.~Poco, H.~T. Vo, J.~Freire, and C.~T. Silva, ``Visual exploration of big spatio-temporal urban data: A study of new york city taxi trips,'' \emph{IEEE transactions on visualization and computer graphics}, vol.~19, no.~12, pp. 2149--2158, 2013.

\bibitem{yuan2011driving}
J.~Yuan, Y.~Zheng, X.~Xie, and G.~Sun, ``Driving with knowledge from the physical world,'' in \emph{Proceedings of the 17th ACM SIGKDD international conference on Knowledge discovery and data mining}, 2011, pp. 316--324.

\bibitem{bbliaojqZhangKDD18deep}
B.~Liao, J.~Zhang, C.~Wu, D.~McIlwraith, T.~Chen, S.~Yang, Y.~Guo, and F.~Wu, ``Deep sequence learning with auxiliary information for traffic prediction,'' in \emph{Proceedings of the 24th ACM SIGKDD International Conference on Knowledge Discovery and Data Mining}.\hskip 1em plus 0.5em minus 0.4em\relax ACM, 2018.

\bibitem{wei2016transfer}
Y.~Wei, Y.~Zheng, and Q.~Yang, ``Transfer knowledge between cities,'' in \emph{Proceedings of the 22nd ACM SIGKDD International Conference on Knowledge Discovery and Data Mining}, 2016, pp. 1905--1914.

\bibitem{zhang2017deep}
J.~Zhang, Y.~Zheng, and D.~Qi, ``Deep spatio-temporal residual networks for citywide crowd flows prediction,'' in \emph{Proceedings of the AAAI Conference on Artificial Intelligence}, vol.~31, no.~1, 2017.

\bibitem{geng2019spatiotemporal}
X.~Geng, Y.~Li, L.~Wang, L.~Zhang, Q.~Yang, J.~Ye, and Y.~Liu, ``Spatiotemporal multi-graph convolution network for ride-hailing demand forecasting,'' in \emph{Proceedings of the AAAI Conference on Artificial Intelligence}, vol.~33, 2019, pp. 3656--3663.

\bibitem{kipf2016semi}
T.~N. Kipf and M.~Welling, ``Semi-supervised classification with graph convolutional networks,'' \emph{arXiv preprint arXiv:1609.02907}, 2016.

\bibitem{velivckovic2017graph}
P.~Veli{\v{c}}kovi{\'c}, G.~Cucurull, A.~Casanova, A.~Romero, P.~Lio, and Y.~Bengio, ``Graph attention networks,'' \emph{arXiv preprint arXiv:1710.10903}, 2017.

\bibitem{yu2018spatio}
B.~Yu, H.~Yin, and Z.~Zhu, ``Spatio-temporal graph convolutional networks: a deep learning framework for traffic forecasting,'' in \emph{Proceedings of the 27th International Joint Conference on Artificial Intelligence}, 2018, pp. 3634--3640.

\bibitem{li2018deeper}
Q.~Li, Z.~Han, and X.-M. Wu, ``Deeper insights into graph convolutional networks for semi-supervised learning,'' in \emph{Proceedings of the AAAI Conference on Artificial Intelligence}, vol.~32, no.~1, 2018.

\bibitem{pan2009survey}
S.~J. Pan and Q.~Yang, ``A survey on transfer learning,'' \emph{IEEE Transactions on knowledge and data engineering}, vol.~22, no.~10, pp. 1345--1359, 2009.

\bibitem{tang2019reinforcement}
X.~Tang, Z.~T. Qin, F.~Zhang, Z.~Wang, Z.~Xu, Y.~Ma, H.~Zhu, and J.~Ye, ``A deep value-network based approach for multi-driver order dispatching,'' in \emph{Proceedings of the 25th ACM SIGKDD International Conference on Knowledge Discovery and Data Mining}, 2019, pp. 1780--1790.

\bibitem{zhou2019multi}
M.~Zhou, J.~Jin, W.~Zhang, Z.~Qin, Y.~Jiao, C.~Wang, G.~Wu, Y.~Yu, and J.~Ye, ``Multi-agent reinforcement learning for order-dispatching via order-vehicle distribution matching,'' in \emph{Proceedings of the 28th ACM International Conference on Information and Knowledge Management}, 2019, pp. 2645--2653.

\bibitem{li2020autost}
T.~Li, J.~Zhang, K.~Bao, Y.~Liang, Y.~Li, and Y.~Zheng, ``Autost: Efficient neural architecture search for spatio-temporal prediction,'' in \emph{Proceedings of the 26th ACM SIGKDD International Conference on Knowledge Discovery \& Data Mining}, 2020, pp. 794--802.

\bibitem{yang2019federated}
Q.~Yang, Y.~Liu, T.~Chen, and Y.~Tong, ``Federated machine learning: Concept and applications,'' \emph{ACM Transactions on Intelligent Systems and Technology (TIST)}, vol.~10, no.~2, pp. 1--19, 2019.

\bibitem{liu2011discovering}
W.~Liu, Y.~Zheng, S.~Chawla, J.~Yuan, and X.~Xing, ``Discovering spatio-temporal causal interactions in traffic data streams,'' in \emph{Proceedings of the 17th ACM SIGKDD international conference on Knowledge discovery and data mining}, 2011, pp. 1010--1018.

\end{thebibliography}

\begin{IEEEbiography}[{\includegraphics[width=1in,height=1.25in,clip,keepaspectratio]{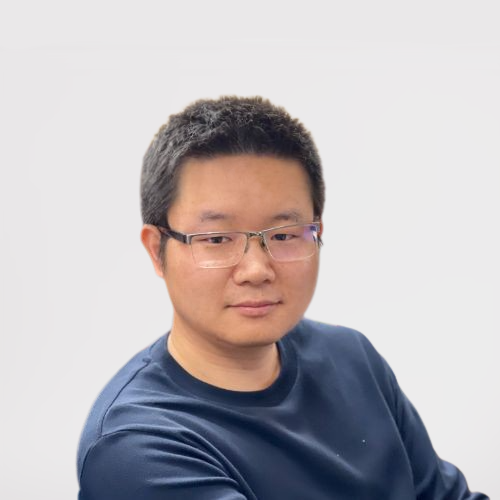}}]{Zhengfei Zheng} received his B.S. and Master degree in transportation engineering from Harbin Institute of Technology, Technion-Israel Institute of Technology, respectively. He obtained his PhD degree in Civil Engineering from the Hong Kong University of Science and Technology (HKUST). Now Zhengfei is a postdoctoral fellow at HKUST. His research interest spans a variety of topics, including emerging smart mobility (ride-sourcing services, bike-sharing, multimodal transportation), AI in transportation (spatiotemporal prediction of traffic states), macroscopic fundamental diagram.
\end{IEEEbiography}
\begin{IEEEbiography}[{\includegraphics[width=1in,height=1.25in,clip,keepaspectratio]{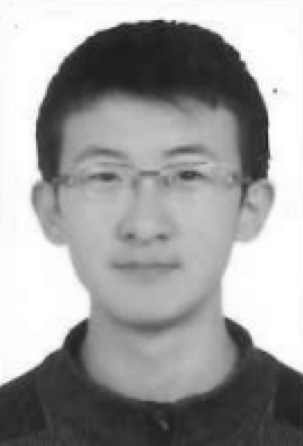}}]{Xu Geng} is a Ph.D. student at the Hong Kong University of Science and Technology, Hong Kong. His research focuses on machine learning, including transfer learning, spatiotemporal learning on smart city applications. He received his bachelor’s degree in computing from the Hong Kong Polytechnic University and master's degree in big data technology from the Hong Kong University of Science and Technology, Hong Kong.
\end{IEEEbiography}
\begin{IEEEbiography}[{\includegraphics[width=1in,height=1.25in,clip,keepaspectratio]{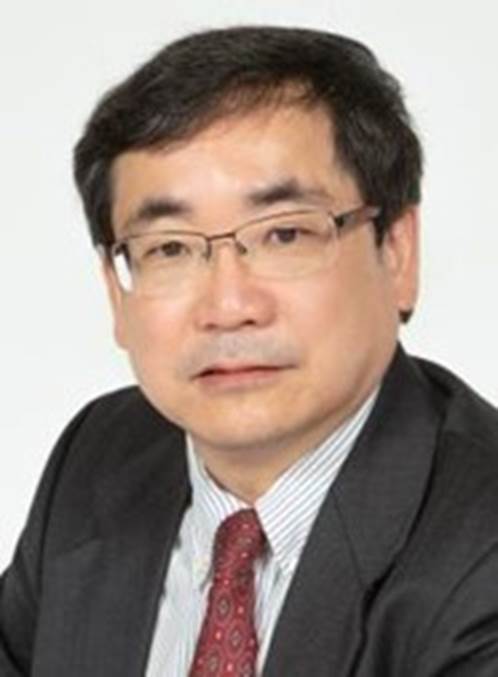}}]{Hai Yang} is currently a Chair Professor in the Department of Civil and Environmental Engineering at the Hong Kong University of Science and Technology, as well as the founding director of the HKUST-DiDi Joint Research Laboratory. He earned his Bachelor's degree from Wuhan University in China and his PhD from Kyoto University in Japan. Prof. Yang specializes in the analysis, modeling, and optimization of transportation systems and transportation economics. He is widely recognized as an active scholar in the field of transportation, with over 400 papers published in SCI/SSCI indexed journals and a Google H-index citation rate of 94 as of March 2024. His research has been published in leading international journals, including Transportation Research, Transportation Science, and Operations Research. Prof. Yang is also a member of the Distinguished Journal Editorial Board of Transportation Research Part B: Methodological, a top journal in the field of transportation.
\end{IEEEbiography}

\newpage
\appendix

\section{Appendix}
\subsection{Summary of Notations}
We summarize all notations used in this paper in Table \ref{tab:notation}. 
\vspace{-0.5cm}
\begin{table}[ht]
    \caption{Summary of Notations}
    \label{tab:notation}
    \centering
    \resizebox{\columnwidth}{!}{
    \begin{tabular}{c|c}
    \toprule
        Notation & Meaning \\
         \midrule
        $c$ & A specific city, e.g. Beijing, Shanghai \\
        $W_c, H_c$ & Width and height of the city $c$ in grids \\
        $r$ & A region, i.e. a 1km$\times$1km grid\\
        $r_c(i, j)$ & The region in city $c$ with grid coordinate $(i, j)$\\
        $\mathcal{R}_c$ & The set of all regions in city $c$\\
        $\tau$ & A timestamp, a 30-minute interval by default\\
        $\tau_c$ & The last valid timestamp of city $c$ \\
        $\mathcal{T}_c$ & The set of valid timestamps of city $c$\\
        $T_c$ & \# valid timestamps of city $c$\\
        $\mathcal{G}_c$ & The set of all taxi GPS points in city $c$\\ 
        $\mathcal{A}_c$ & The set of all taxis in city $c$\\
        $tr_a$ & The trajectory of taxi $a$\\
        $(pos, t, occ)$ & A taxi GPS point\\
        $POI_c$ & The set of POIs in city $c$\\
        $(pos_{poi}, cat_{poi})$ & The position and category of a POI\\
        $\mathbf{x}_c$ & A spatio-temporal tensor for city $c$\\
        $\mathbf{x}_c(i, j, \tau)$ & The value at region $r_c(i, j)$ at $\tau$  \\
        $\mathbf{x}_c(\tau)$ & The spatio-temporal values of $\mathcal{R}_c$ at time $\tau$\\
        $seg = (pos_0, pos_1)$ & A road segment and its two endpoints\\
        $SEG_c$ & The set of road segments in city $c$\\
        $sp_{seg}$ & The sequence of history traffic speeds of $seg$\\
    \bottomrule
    \end{tabular}}

\end{table}
\subsection{Details of Meteorological Dataset}
The weather data matrix $\mathbf{x}_c^{mtr}\in \mathbb{R}^{T_c/2\times 49}$ consists of $T_c/2$ rows and 49 columns. Each dimension in the 49-dimensional vector $\mathbf{x}_c^{mtr}(\tau)$ represents a specific aspect of the weather data. The 49 dimensions of the vector $\mathbf{x}_c^{mtr}(\tau)$ have the following meanings:
\begin{itemize}
    \item 0: Temperature (°C);
    \item 1: Atmospheric pressure (mmHg);
    \item 2: Sea level pressure (mmHg);
    \item 3: Relative humidity (\%);
    \item 4: Wind speed (m/s);
    \item 5: Horizontal visibility (km);
    \item 6: Dew point temperature (°C);
    \item 7-24: Multi-hot encodings of whether phenomena. Each index represents \textit{fog, patches fog, partial fog, mist, haze, widespread dust, light drizzle, rain, light rain, shower, light shower, in the vicinity shower, heavy shower, thunderstorm, light thunderstorm, in the vicinity thunderstorm, heavy thunderstorm}, and \textit{no special weather phenomena}, respectively;
    \item 25-42: One-hot encoding of wind direction. Each index represents \textit{no wind, variable direction, E, E-NE, E-SE, W, W-NW, W-SW, S, S-SE, S-SW, SE, SW, N, N-NE, N-NW, NE}, and \textit{NW}, respectively;
    \item 43-48: One-hot encoding of wind covering. Each index represents \textit{no significant clouds, cumulonimbus clouds, few clouds, scattered clouds, broken clouds}, and \textit{overcast}, respectively. 
\end{itemize}
\subsection{Details of Machine Learning Experiments}
 We would like to emphasize that the settings for our machine learning experiments were chosen to ensure that all models converge, rather than to achieve state-of-the-art results. Below are the detailed settings used in our experiments.
\subsubsection{Taxi Service Prediction}
 For our CNN and LSTM models, we trained them using the following settings:
\begin{itemize}
    \item Learning rate: 1e-4
    \item Weight decay: 0
    \item Optimization algorithm: Adam
    \item Number of epochs: 75
    \item Batch size: 64 for CNN, variable batch sizes (Beijing 4, Shanghai 8, Shenzhen/Chongqing 16, Xi'an/Chengdu 32) for LSTM
    \item Model selection: choose the model with the lowest validation error for testing
    \item Weather data usage: following the method in \cite{zhang2017deep}, weather data are used in training the model
    \item Number of independent runs: 5
    \item Evaluation metric: Root Mean Squared Error (RMSE) and Mean Absolute Error (MAE)
\end{itemize}

\subsubsection{Inter-city Transfer Learning}
To fine-tune our model, we utilized an Adam optimizer with a learning rate of 5e-5 and weight decay of 0. This process was carried out over 100 epochs on the source city dataset. The model with the lowest validation error on the target city dataset was selected for testing. We also implemented the RegionTrans method, with a weight of 0.01 assigned to feature consistency.


\subsection{Hardware Configurations and Data-sharing Approaches}
The experiments presented in this paper were conducted using Python 3.6 and PyTorch 1.6.0 in a computing environment equipped with Tesla V100 GPUs, which are affiliated with our computing platform.

To facilitate the development of smart city applications in various domains, such as transportation, healthcare, and finance, we are currently constructing an open platform that provides high-performance computing resources, extensive smart city data, including our CityNet dataset, and cutting-edge algorithmic paradigms. Researchers are welcome to utilize our platform to train their own machine learning models on CityNet, which is available for free on our cluster equipped with GPUs and RDMA.

We have opened our platform to the local academic community and global researchers upon request, following a thorough assessment of data privacy issues. Some of the original data utilized in this study were obtained from publicly available sources, while others were retrieved from the HKUST-DiDi Joint Research Laboratory. To ensure the protection of personal information, some of the data can be made available upon request after undergoing a process of desensitization. Our goal is to promote collaboration and advance research in the field of urban computing and smart cities.


\end{document}